\documentclass[sigconf]{acmart}
\AtBeginDocument{%
  }

\copyrightyear{2026}
\acmYear{2026}
\setcopyright{cc}
\setcctype{by}
\acmConference[WWW '26]{Proceedings of the ACM Web Conference 2026}{April 13--17, 2026}{Dubai, United Arab Emirates}
\acmBooktitle{Proceedings of the ACM Web Conference 2026 (WWW '26), April 13--17, 2026, Dubai, United Arab Emirates}
\acmPrice{}
\acmDOI{10.1145/3774904.3793007}
\acmISBN{979-8-4007-2307-0/2026/04}
\usepackage{makecell}
\usepackage{multirow}
\usepackage{siunitx}
\usepackage{stfloats}

\usepackage{subcaption}

\usepackage{soul}

\usepackage{enumitem}
\usepackage{cleveref}

\renewcommand{\arraystretch}{1.1}

\begin{document}

%%
%% The "title" command has an optional parameter,
%% allowing the author to define a "short title" to be used in page headers.
\title[Cross-Cultural Evaluation of Vision–Language Models for Hateful Meme Detection]%
{From Native Memes to Global Moderation: Cross-Cultural Evaluation of Vision–Language Models for Hateful Meme Detection}

\author{Mo Wang}
\email{momow818@gmail.com}
\orcid{0009-0005-1268-6411}
\affiliation{%
  \institution{Macquarie University}
  \city{Sydney}
  \country{Australia}
}

\author{Kaixuan Ren}
\orcid{0009-0006-5547-5240}
\email{kaixuan.ren@mq.edu.au}
\affiliation{%
  \institution{Macquarie University}
  \city{Sydney}
  \country{Australia}
}

\author{Pratik Jalan}
\orcid{0009-0005-8348-0509}
\email{pratikjalan9@gmail.com}
\affiliation{%
  \institution{Macquarie University}
  \city{Sydney}
  \country{Australia}
}

\author{Ahmed Ashraf}
\orcid{0009-0001-6806-3038}
\email{ahme.ashraf@nu.edu.eg}
\affiliation{%
  \institution{Nile University}
  \city{Giza}
  \country{Egypt}
}

\author{Tuong Vy Vu}
\orcid{0009-0004-6335-2027}
\email{tuongvy.vu@students.mq.edu.au}
\affiliation{%
  \institution{Macquarie University}
  \city{Sydney}
  \country{Australia}
}

\author{Rahul Seetharaman}
\orcid{0000-0001-6171-4814}
\email{rahulseetharaman@gmail.com}
\affiliation{%
  \institution{University of Massachusetts Amherst}
  \city{Amherst}
  \country{United States}
}

\author{Shah Nawaz}
\orcid{0000-0002-7715-4409}
\email{shah.nawaz@jku.at}
\affiliation{%
  \institution{Johannes Kepler University}
  \city{Linz}
  \country{Austria}
}

\author{Usman Naseem}
\orcid{0000-0003-0191-7171}
\email{usman.naseem@mq.edu.au}
\affiliation{%
  \institution{Macquarie University}
  \city{Sydney}
  \country{Australia}
}

%%
%% By default, the full list of authors will be used in the page
%% headers. Often, this list is too long, and will overlap
%% other information printed in the page headers. This command allows
%% the author to define a more concise list
%% of authors' names for this purpose.
% \renewcommand{\shortauthors}{Trovato et al.}
\renewcommand{\shortauthors}{Mo Wang et al.}

%%
%% The abstract is a short summary of the work to be presented in the
%% article.
\begin{abstract}
Cultural context profoundly shapes how people interpret online content, yet vision–language models (VLMs) remain predominantly trained through Western or English-centric lenses. This limits their fairness and cross-cultural robustness in tasks like hateful meme detection. We introduce a systematic evaluation framework designed to diagnose and quantify the cross-cultural robustness of state-of-the-art VLMs across multilingual meme datasets, analyzing three axes: (i) learning strategy (zero-shot vs. one-shot), (ii) prompting language (native vs. English), and (iii) translation effects on meaning and detection. Results show that the common ``translate-then-detect'' approach deteriorate performance, while culturally aligned interventions — native-language prompting and one-shot learning — significantly enhance detection. Our findings reveal systematic convergence toward Western safety norms and provide actionable strategies to mitigate such bias, guiding the design of globally robust multimodal moderation systems.

\end{abstract}

%%
%% The code below is generated by the tool at http://dl.acm.org/ccs.cfm.
%% Please copy and paste the code instead of the example below.
%%
% \ccsdesc[500]{Information systems~Multimedia retrieval}


%%
%% Keywords. The author(s) should pick words that accurately describe
%% the work being presented. Separate the keywords with commas.

% \received{20 February 2007}
% \received[revised]{12 March 2009}
% \received[accepted]{5 June 2009}

%%
%% This command processes the author and affiliation and title
%% information and builds the first part of the formatted document.
\maketitle

\begin{figure*}[ht!]
    \centering
    \includegraphics[width=0.85\textwidth]{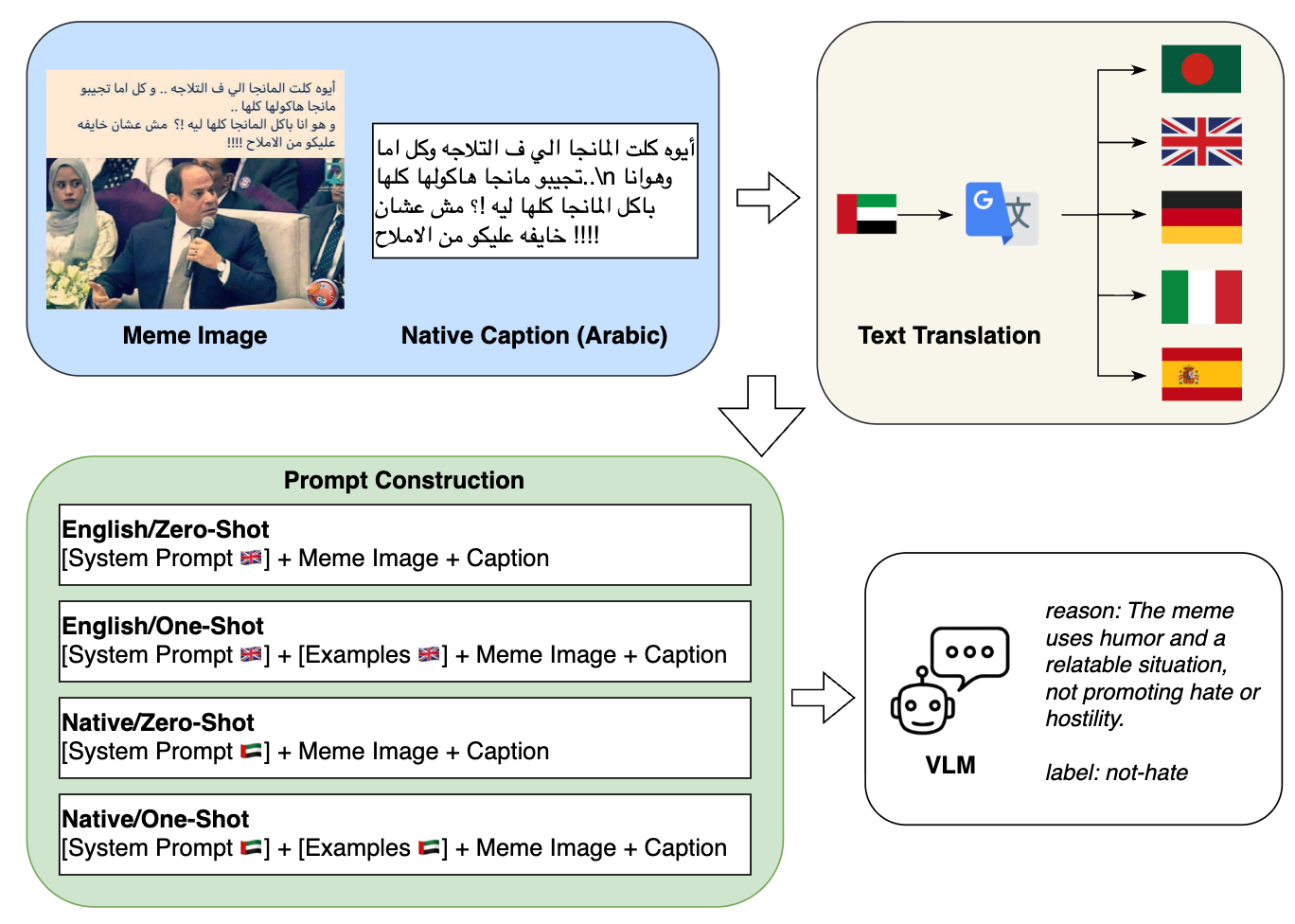}
    \Description{Overall workflow of our multilingual hateful meme detection evaluation.}
    \caption{
        Overall workflow of our multilingual hateful meme detection evaluation. Starting from native memes (e.g., Arabic), we apply optional machine translation (MT) into multiple target languages, followed by the construction of different prompt types (English/Native × Zero-Shot/One-Shot). These prompts are then fed into VLMs, whose predictions and explanations are used to assess cross-lingual cultural robustness under different interaction strategies.
    }
    \label{fig:main_framework}
    \vspace{-3mm}
\end{figure*}

\section{Introduction}

Cultural background plays a pivotal role in shaping individuals' perceptions and interpretations of digital content. As highlighted by~\citep{bui-etal-2025-multi3hate}, an internet meme can provoke vastly different interpretations across diverse cultural groups. This phenomenon presents significant challenges for platforms striving to moderate hateful content. Recent research has shown that large VLMs often reflect Western or English-centric cultural norms in both their representations and predictions~\citep{ananthram2025perspectivelanguage, liu2025culturevlm}. For example,~\citet{ananthram2025perspectivelanguage} reveal how language choice can influence the cultural bias of image understanding models, while~\citet{liu2025culturevlm} systematically benchmark VLMs across over $100$ countries, uncovering significant cross-cultural disparities. Such biases can marginalize non-English cultures, posing a critical issue for fair and robust evaluation of global content moderation systems.

Multimodal learning remains heavily skewed towards English, with influential resources such as the Hateful Memes dataset~\citep{kiela2021hatefulmemeschallengedetecting} defining much of the research landscape. Pioneering efforts such as the Multi3Hate~\citep{bui-etal-2025-multi3hate} have made substantial progress by introducing multilingual translations. While recent studies have expanded to specific cultural or topical domains, such as political conflicts~\citep{polimeme2025} or gender-targeted abuse~\cite{chen2024unveiling}, these efforts remain isolated to single contexts. However, most work still originates from English-centric content, inadequately capturing the  humor, symbolic cues, and contextual nuances present in memes from non-English communities. This creates a critical gap, as the performance of moderation models in  native contexts is rarely tested, necessitating new analytical approaches that can rigorously evaluate their cross-cultural capabilities.

While recent studies have expanded to specific cultural or topical domains such as political conflicts~\cite{polimeme2025} or gender-targeted abuse~\cite{chen2024unveiling}, these efforts remain isolated to single contexts.
In contrast, our study evaluates VLMs on \textbf{native, culturally grounded} memes spanning six distinct linguistic ecosystems. This enables us to probe model weaknesses that cannot be uncovered in translation-based benchmarks.

To address this gap, this paper introduces a systematic, multifaceted evaluation framework to probe state-of-the-art VLMs across diverse cultural contexts in hateful meme detection task. We leverage existing multicultural resources to investigate which strategies may strengthen or weaken detection performance across diverse linguistic contexts. Specifically, our study dissects model behavior across three critical axes: (i) learning strategies (zero-shot vs. one-shot learning), (ii) prompting strategies (native-language vs. English-centric prompts), and (iii) the impact of translating meme captions across multiple non-English languages. This leads to our central research question: To what extent can VLMs achieve robust and transferable hateful meme detection, and which interventions are most effective at bridging the cultural understanding gap?

Our contributions are threefold.

\begin{itemize}
    \item We present a multi-dimensional framework for multilingual hateful meme detection, assessing VLMs across learning strategies, prompts, and content representations.
    \item We benchmark models to identify which strategies strengthen or weaken detection performance, providing systematic empirical evidence that ‘translate-then-detect’ underperforms while native-language prompting and one-shot learning are highly effective.
    \item We provide evidence of inherent Western-centric bias in VLMs and demonstrate that culturally aligned interventions can effectively mitigate these biases, improving cross-cultural robustness.
\end{itemize}

\section{Related Work}

% The detection of harmful content has evolved from simple textual analysis to complex, multimodal, and multilingual challenges. Internet memes, as potent carriers of cultural expression where humor and sarcasm vary widely across groups~\citep{shifman-2013-memes}, present a particularly difficult challenge for this task. Our work is situated at the intersection of three key areas: (1) the development of multilingual and multimodal datasets, (2) the study of cross-cultural annotation and bias, and (3) the evaluation of cultural robustness in VLMs.
The detection of harmful content has evolved from simple textual analysis to complex, multimodal, and multilingual challenges. Internet memes, as potent carriers of cultural expression where humor and sarcasm vary widely across groups~\citep{shifman-2013-memes}, present a particularly difficult challenge for this task. Our work is situated at the intersection of three key areas: (1) the development of multilingual and multimodal benchmarks, (2) the study of cross-cultural annotation and bias, and (3) the evaluation of cultural robustness in VLMs.

\subsection{Multilingual and Multimodal Benchmarks}

Recent research has expanded beyond monolingual settings. In the textual domain, benchmarks like XHate-$999$~\citep{glavas-etal-2020-xhate} have supported the creation of cross-lingual models. However, the multimodal domain remains heavily skewed towards English. Benchmark datasets such as Hateful Memes~\citep{kiela2021hatefulmemeschallengedetecting} and CrisisHateMM~\citep{bhandari-etal-2023-crisishatemm}, while crucial, are curated around English-centric cultural contexts, limiting their utility for evaluating models across diverse cultural settings.

While efforts like MUTE~\citep{hossain-etal-2022-mute} have provided valuable non-English resources, and other works have curated native datasets for low-resource languages like Nepali~\cite{thapa2025nememe} or specific domains like health misinformation~\cite{naseem2023vaccine}, the field still lacks large-scale datasets with a parallel structure across multiple languages and cultures. For instance, Multi3Hate~\citep{bui-etal-2025-multi3hate} is a key multilingual benchmark, yet its content originates from English sources and is subsequently translated. This methodology, while pioneering, inherently limits the capture of native, organically sourced cultural content. Our study addresses this gap not by creating a new dataset, but by pioneering a multi-faceted evaluation methodology that leverages existing native, organically sourced memes to probe model capabilities far beyond what standard, translation-based multilingual benchmarks allow.

\subsection{Cross-Cultural Analysis in Hate Speech}

There is growing recognition that ``hate speech'' is not a monolith; its interpretation is shaped by cultural norms. CREHate~\citep{lee-etal-2024-crehate} was a pioneering text-based study that incorporated multicultural perspectives, revealing a systematic alignment between large language model predictions and American cultural norms. However, this work was limited to text and English-language data. Even in multimodal efforts such as Multi3Hate~\citep{bui-etal-2025-multi3hate}, the diversity of annotations is constrained because the source content itself originates from English-centric internet cultures.

This leaves a critical question unanswered: how do state-of-the-art VLMs interpret harmful content that is curated within and for non-English cultural contexts? Our work directly tackles this question by evaluating models on native memes from six diverse language communities, allowing us to analyze how model judgments align with or deviate from the content's original cultural grounding.

\subsection{Evaluating Cultural Robustness in VLMS}

Broad frameworks, such as GOAT-Bench~\citep{khanna-etal-2024-goatbench}, have been instrumental in demonstrating that powerful models like GPT-4V~\citep{openai-2023-gpt4v} and CogVLM~\citep{wang-etal-2024-cogvlm} struggle to detect implicit, multimodal harmful content. 
%Yet, their reliance on English data constrains their utility for cross-cultural assessment.
However, their reliance on English data limits their applicability for robust cross-cultural assessment.
Furthermore, studies on models like CLIP~\citep{radford-etal-2021-clip} have revealed that significant performance deterioration in culturally diverse settings, with cross-lingual evaluations~\citep{garcia2020multilingualviewunsupervisedmachine} showing a failure to generalize beyond the Western norms embedded in the training data. While cultural awareness benchmarks exist for other tasks, such as VQA~\citep{romero2024cvqa}, a framework for evaluating content with a focus on mitigation strategies remains underdeveloped.
These studies effectively diagnose the problem of cultural bias. On the other hand, our work moves from diagnosis to intervention. We take a step further by systematically investigating which strategies such as the choice between native and English prompts, the application of zero-shot versus one-shot learning, and the utility of cross-language translation pipelines can actively improve or deteriorate a VLM's cultural robustness in the critical task of hateful meme detection. This focus on actionable strategies is our key contribution to the field of VLM safety evaluation.

% \begin{table*}[t]
% \centering
% \caption{Statistics of the six datasets used for hateful meme detection. We report total samples and the original label distributions from the source datasets.}
% \label{tab:hateful_data_stats}
% \scriptsize
% \setlength{\tabcolsep}{4pt}
% \renewcommand{\arraystretch}{1.1}

% \scalebox{0.99}{
% \begin{tabular}{@{}l l r p{0.52\textwidth}@{}}
% \toprule
% \textbf{Language} & \textbf{Dataset} & \textbf{Total} & \textbf{Label Breakdown} \\
% \midrule
% Arabic & Prop2Hate~\citep{alam2024propaganda} & 3,061 &
% \texttt{hate}: 398 (13.0\%), \texttt{not-hate}: 2,663 (87.0\%) \\

% Bengali & BHM~\citep{hossain-etal-2024-deciphering} & 6,852 &
% \texttt{hate}: 2,557 (37.3\%), \texttt{non-hate}: 4,295 (62.7\%) \\

% English & HateMeme~\citep{kiela2021hatefulmemeschallengedetecting} & 5,029 &
% \texttt{hateful}: 1,810 (36.0\%), \texttt{non-hateful}: 3,219 (64.0\%) \\

% German & GerMemeHate~\citep{bui-etal-2025-multi3hate} & 179 &
% \texttt{hate}: 104 (58.1\%), \texttt{non-hate}: 75 (41.9\%) \\

% Italian & DANKMEMES~\citep{giorgi2021dankmemes} & 1,000 &
% \texttt{hate}: 500 (50.0\%), \texttt{non-hate}: 500 (50.0\%) \\

% Spanish & DIMEMEX~\citep{GarcaHidalgo2024DIMEMEX2024CA} & 2,263 &
% \texttt{hate\_speech}: 386 (17.1\%), \texttt{inappropriate\_content}: 472 (20.9\%), \texttt{neither}: 1,405 (62.1\%) \\
% \bottomrule
% \end{tabular}
% }
% \vspace{-1mm}
% \end{table*}

\section{Benchmarking}
\label{sec:benchmarking}

\subsection{Overview and Evaluation Pipeline}
\label{sec:pipeline}

We design the experimental setup to rigorously evaluate the cross-lingual and cross-cultural capabilities of VLMs on hateful meme detection task. 
To systematically assess model robustness, we adopt a multidimensional benchmarking design that jointly varies model type, learning paradigm, prompt language, and content representation (native vs. translated). 
The complete workflow is illustrated in Figure~\ref{fig:main_framework}.

The evaluation proceeds through three primary stages:
\begin{enumerate}
    \item Data Acquisition and Cultural Grounding: We select six datasets containing locally created memes within Arabic, Bengali, English, German, Italian, and Spanish cultures, ensuring authentic cultural context (Section \ref{sec:datasets_cultural_scope}).
    \item Content Representation: We create parallel sets by translating meme captions (text only) into all other target languages using Google Translate. Crucially, the original image and any image-embedded text remain unaltered (Section \ref{sec:translation_protocol}).
    \item VLM Probing: We evaluate models under a $2\times2$ prompting grid: Native/English Prompt $\times$ Zero-shot/One-shot learning (Section \ref{sec:prompting_strategy}).
\end{enumerate}

An overview of this design is provided in Table~\ref{tab:benchmark_summary}, which summarizes the dimensions and their corresponding purposes within our evaluation.

\subsection{Datasets and Cultural Scope}
\label{sec:datasets_cultural_scope}
Six publicly available datasets spanning diverse languages and social contexts are used (see Table~\ref{tab:hateful_data_stats} for statistics): Prop2Hate (Arabic) \citep{alam2024propaganda}, BHM (Bengali) \citep{hossain-etal-2024-deciphering}, HateMeme (English) \citep{kiela2021hatefulmemeschallengedetecting}, GerMemeHate (German) \citep{bui-etal-2025-multi3hate}, DANKMEMES (Italian) \citep{giorgi2021dankmemes}, and DIMEMEX (Spanish) \citep{GarcaHidalgo2024DIMEMEX2024CA}.

\begin{table}[t]
\centering
\caption{Statistics of the six datasets used for hateful meme detection task. We report total samples and the original label distributions from the source datasets.}
\label{tab:hateful_data_stats}
\Description{A table listing six datasets by language, showing total sample counts and the original label distributions for each dataset.}
\footnotesize
\setlength{\tabcolsep}{3pt}
\renewcommand{\arraystretch}{1.1}

\begin{tabular}{@{}p{0.12\columnwidth} p{0.20\columnwidth} r p{0.56\columnwidth}@{}}
\toprule
\textbf{Language} & \textbf{Dataset} & \textbf{Total} & \textbf{Label Breakdown} \\
\midrule
Arabic & Prop2Hate~\citep{alam2024propaganda} & 3,061 &
\texttt{hate}: 398 (13.0\%), \texttt{not-hate}: 2,663 (87.0\%) \\
Bengali & BHM~\citep{hossain-etal-2024-deciphering} & 6,852 &
\texttt{hate}: 2,557 (37.3\%), \texttt{non-hate}: 4,295 (62.7\%) \\
English & HateMeme~\citep{kiela2021hatefulmemeschallengedetecting} & 5,029 &
\texttt{hateful}: 1,810 (36.0\%), \texttt{non\allowbreak-hateful}: 3,219 (64.0\%) \\
German & GerMemeHate~\citep{bui-etal-2025-multi3hate} & 179 &
\texttt{hate}: 104 (58.1\%), \texttt{non-hate}: 75 (41.9\%) \\
Italian & DANKMEMES~\citep{giorgi2021dankmemes} & 1,000 &
\texttt{hate}: 500 (50.0\%), \texttt{non\allowbreak-hate}: 500 (50.0\%) \\
Spanish & DIMEMEX~\citep{GarcaHidalgo2024DIMEMEX2024CA} & 2,263 &
\texttt{hate\_speech}: 386 (17.1\%), \texttt{inappropriate\allowbreak\_content}: 472 (20.9\%), \texttt{neither}: 1,405 (62.1\%) \\
\bottomrule
\end{tabular}
\vspace{-1mm}
\end{table}

\paragraph{Cultural Grounding.}
We emphasize that these datasets represent distinct cultural ecosystems, not merely different languages. For example, Arabic (Prop2Hate) focuses on region-specific political propaganda and religious symbolism; Bengali (BHM) captures South Asian code-mixing and local political humor; and Spanish (DIMEMEX) centers on Mexican Spanish multimodal abuse, including machismo and classism. This native cultural context is essential for validating the cross-cultural robustness of VLMs.

\subsection{Models}
Our evaluation includes both general-purpose VLMs and task-specific models for hateful content detection.  

\paragraph{General-Purpose VLMs.} These models are evaluated primarily in zero-shot settings to measure their intrinsic capabilities on culturally nuanced content:  
\begin{itemize}
    \item \textbf{Gemini-2.5-Flash}~\citep{comanici2025gemini25}: Efficient MoE architecture optimized for long-context reasoning with strong multimodal performance.
    \item \textbf{GPT-4o-Mini}~\citep{openai-2023-gpt4v}: Low-latency, cost-efficient model with improved tokenizer and 128K context window; excels in MMLU and multimodal benchmarks.
    \item \textbf{CogVLM2}~\citep{hong2024cogvlm2}: High-resolution image handling without resizing; English and Chinese 19B chat-tuned variants evaluated.
    \item \textbf{Qwen 2.5-VL}~\citep{wang2024qwen2vlenhancingvisionlanguagemodels}: Supports images up to 4K pixels; 7B instruction-tuned version used.
    \item \textbf{InstructBLIP}~\citep{dai2023instructblip}: Instruction-tuned on Vicuna-7B, extracts instruction-relevant visual features for general-purpose multimodal tasks.
    \item \textbf{LLaMA-4-Maverick}~\citep{touvron2023llama}: A multimodal model from the LLaMA family, representing state-of-the-art open-foundation architectures.
\end{itemize}

\paragraph{Task-Specific Models.} 
These models are designed specifically for hateful or harmful meme detection and are fine-tuned separately on each of the six native meme datasets for cross-lingual evaluation (see Appendix \ref{sec:appendix_model_details} for training corpora details):
\begin{itemize}
    \item \textbf{Pro-Cap}~\citep{cao2023procap}: Two-stage framework generating detailed captions via a frozen VLM, then feeding combined text into a text-only classifier.
    \item \textbf{PromptHate}~\citep{cao2023promptingmultimodalhatefulmeme}: Reformulates detection as a prompt-based task; uses structured textual prompts for pretrained language models.
\end{itemize}

\subsection{Benchmark and Task}
The hateful meme detection task is formulated as a multi-class classification problem. Given a meme’s image and caption, the model predicts one of the original labels provided by each dataset (e.g., \texttt{hate}, \texttt{not-hate}, or \texttt{inappropriate\_content}). By retaining native label sets, we ensure that the evaluation remains faithful to the cultural and linguistic context of the original data.

\subsection{Prompting Strategy and In-Context Examples}
\label{sec:prompting_strategy}
We employ a $2\times2$ framework: learning paradigm (\textbf{zero-shot vs. one-shot}) $\times$ prompt language (\textbf{English vs. native}). Zero-shot models receive only a task definition; one-shot models receive a single gold-standard example per label, which is selected by native speakers to be a culturally representative exemplar (see Appendix \ref{sec:appendix_prompts} for details). This setup measures intrinsic knowledge and cross-cultural adaptability.

\subsection{Translated Dataset}
We construct parallel datasets via Google Translate, translating each caption into all other target languages while keeping images unchanged. This isolates the effect of language in cross-cultural evaluation.

\begin{table}[!t]
\caption{Overview of the multidimensional benchmarking design.}
\label{tab:benchmark_summary}
\centering
\footnotesize
\resizebox{\columnwidth}{!}{
\begin{tabular}{l l l}
\toprule
\textbf{Dimension} & \textbf{Variants} & \textbf{Purpose} \\
\midrule
Model             & 7 general, 2 specialized & General vs.\ task-specific \\
Learning Paradigm & Zero-shot, One-shot      & In-context adaptability   \\
Prompt Language   & English, Native          & Cultural alignment         \\
Dataset           & 6 languages              & Cross-cultural coverage    \\
\bottomrule
\end{tabular}
}
\vspace{-1mm}
\end{table}

\subsection{Translated Content Protocol}
\label{sec:translation_protocol}
We construct fully parallel content representations via Google Translate (v3 API), translating each meme's caption text only into all other target languages (e.g., Arabic caption $\rightarrow$ Bengali, English, etc.). Crucially, the original meme image, including any text visually embedded in the image, remains unchanged and untranslated. This design allows us to isolate the effect of the caption's linguistic transformation on cross-cultural evaluation, mimicking the common ``translate-then-detect'' pipeline while avoiding the high noise and instability introduced by OCR-based image reconstruction.

\subsection{Evaluation Metrics}
We report Accuracy and Macro-F1 as primary metrics, computed only over valid predictions that match the label schema. The proportion of invalid outputs (e.g., format violations) was consistently very low across models and languages ($<0.5\%$). Therefore, excluding these cases is unlikely to affect the reported results.

\begin{figure*}[t]
\centering
\setlength{\tabcolsep}{2pt}

% ---------------- Row 1: Zero-shot (4 models) ----------------
% \textbf{Zero-shot Performance Across Models}\\[3pt]

\begin{tabular}{cccc}
\begin{minipage}{0.23\linewidth}
\centering
\includegraphics[width=\linewidth]{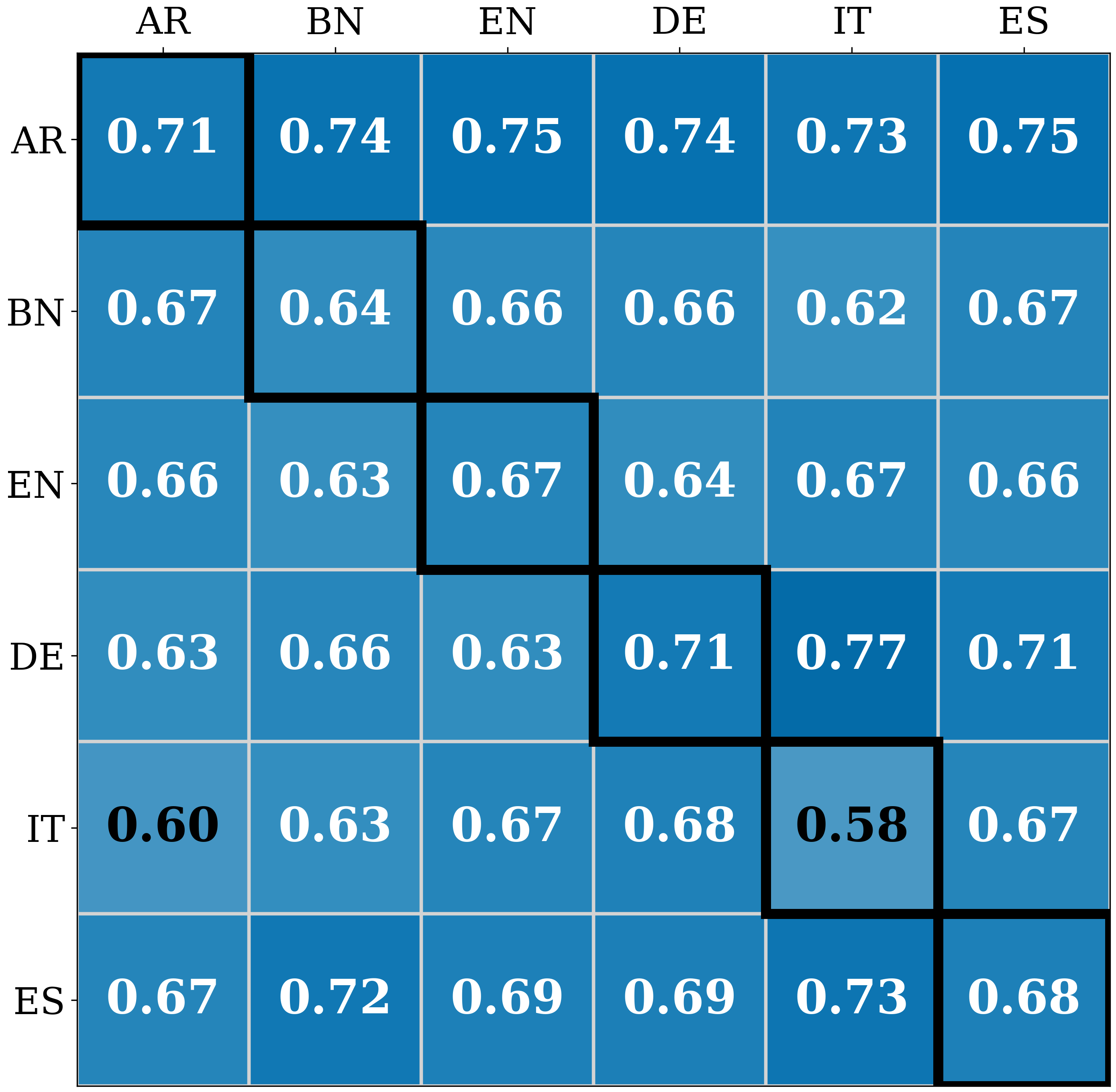}
\vspace{2pt}
{\scriptsize Gemini}
\end{minipage} &
\begin{minipage}{0.23\linewidth}
\centering
\includegraphics[width=\linewidth]{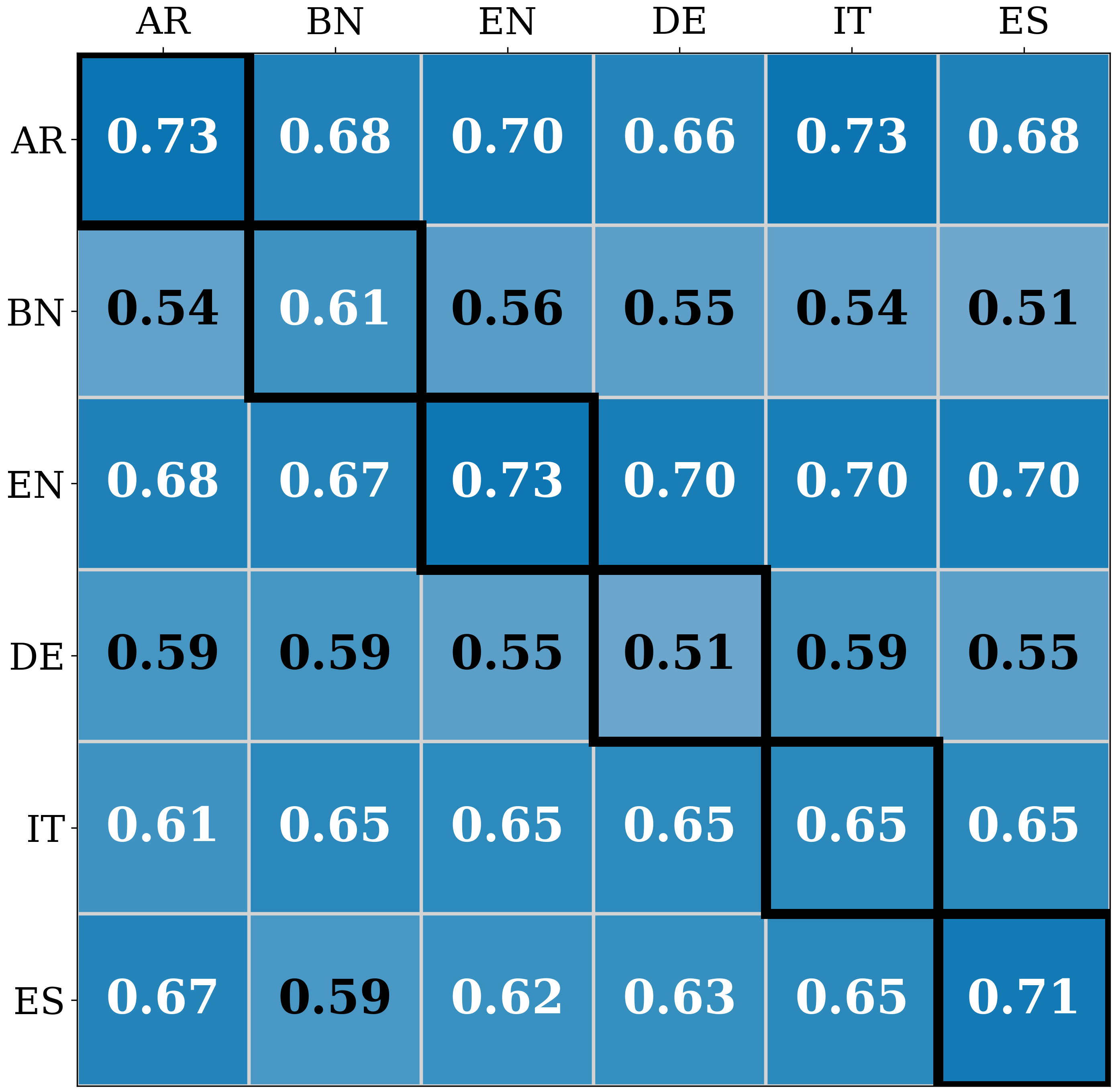}
\vspace{2pt}
{\scriptsize GPT-4o-Mini}
\end{minipage} &
\begin{minipage}{0.23\linewidth}
\centering
\includegraphics[width=\linewidth]{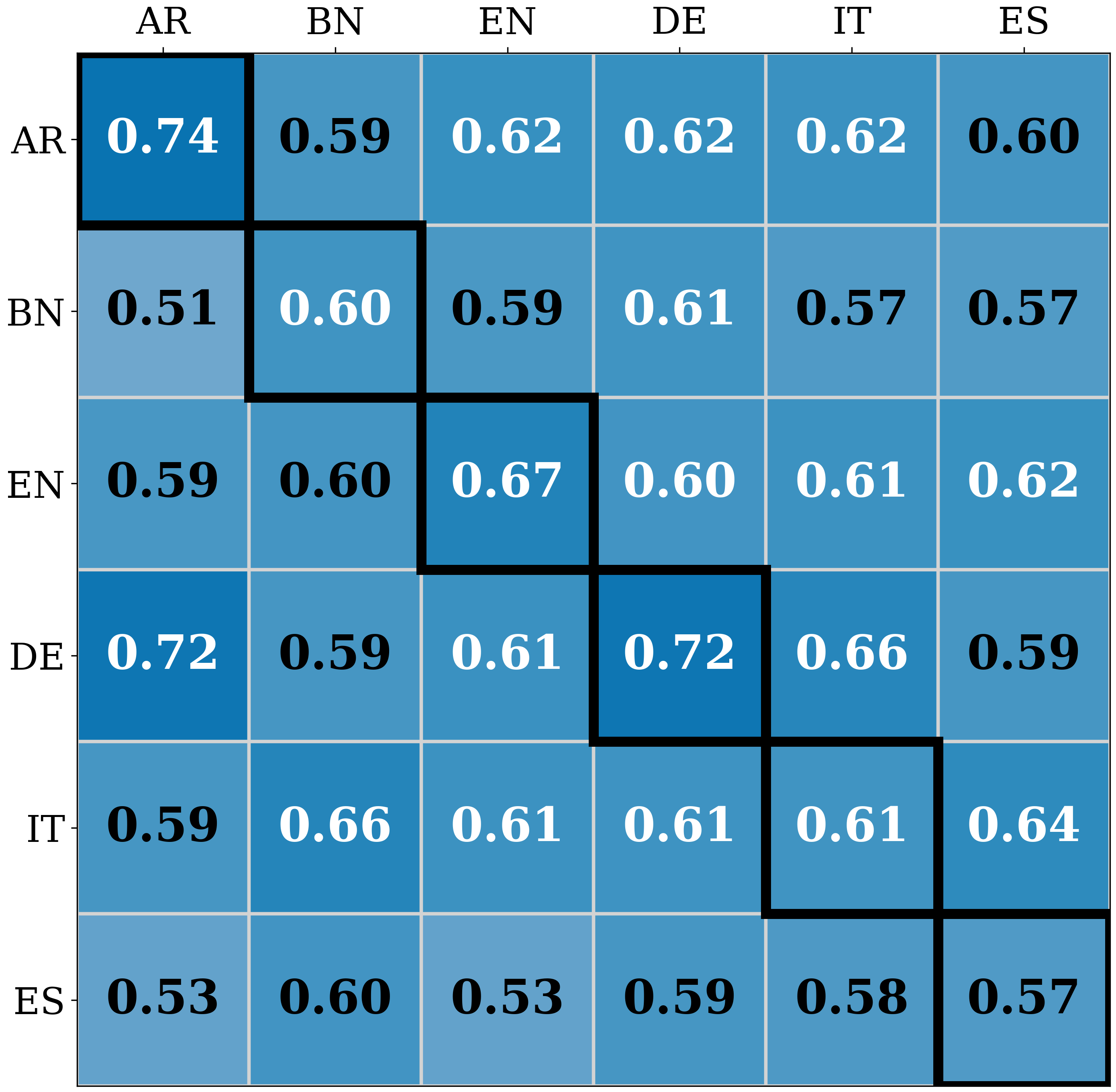}
\vspace{2pt}
{\scriptsize LLaMA-4}
\end{minipage} &
\begin{minipage}{0.23\linewidth}
\centering
\includegraphics[width=\linewidth]{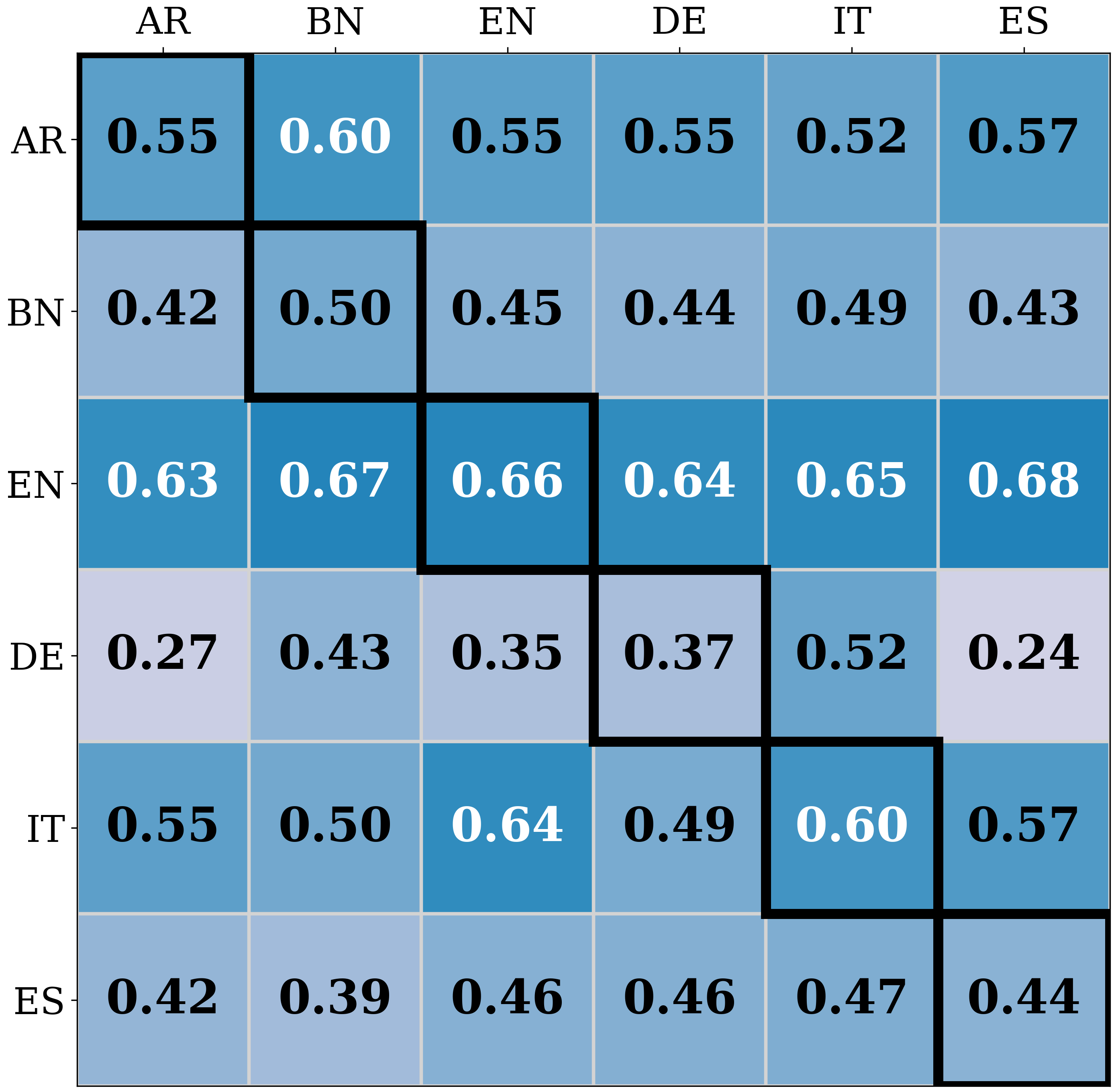}
\vspace{2pt}
{\scriptsize Qwen-VL}
\end{minipage}
\end{tabular}

\vspace{5pt}

% ---------------- Row 2: Zero-shot (remaining 3 models) ----------------
\begin{tabular}{ccc}
\begin{minipage}{0.23\linewidth}
\centering
\includegraphics[width=\linewidth]{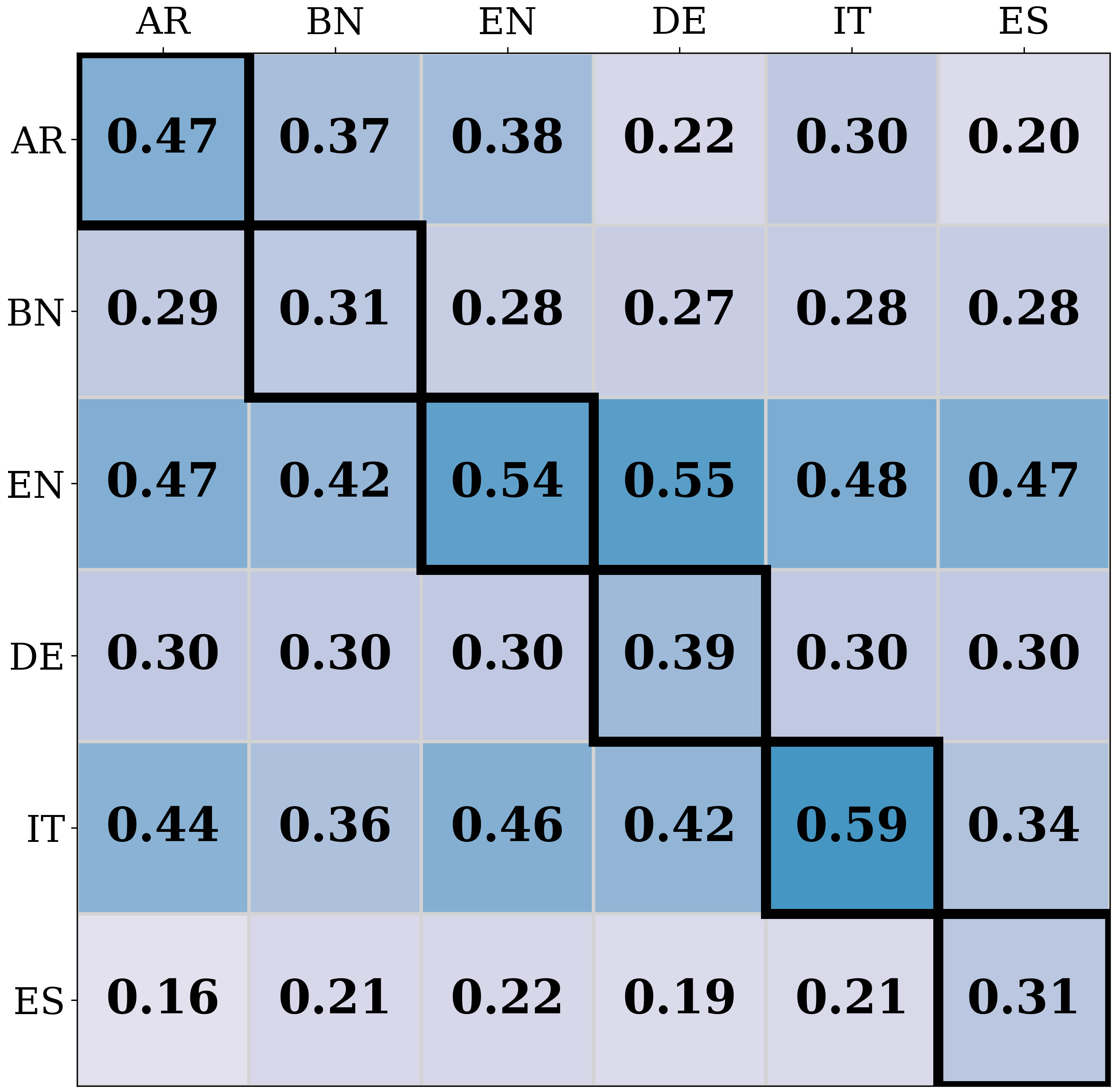}
\vspace{2pt}
{\scriptsize CogVLM2-ZH}
\end{minipage} &
\begin{minipage}{0.23\linewidth}
\centering
\includegraphics[width=\linewidth]{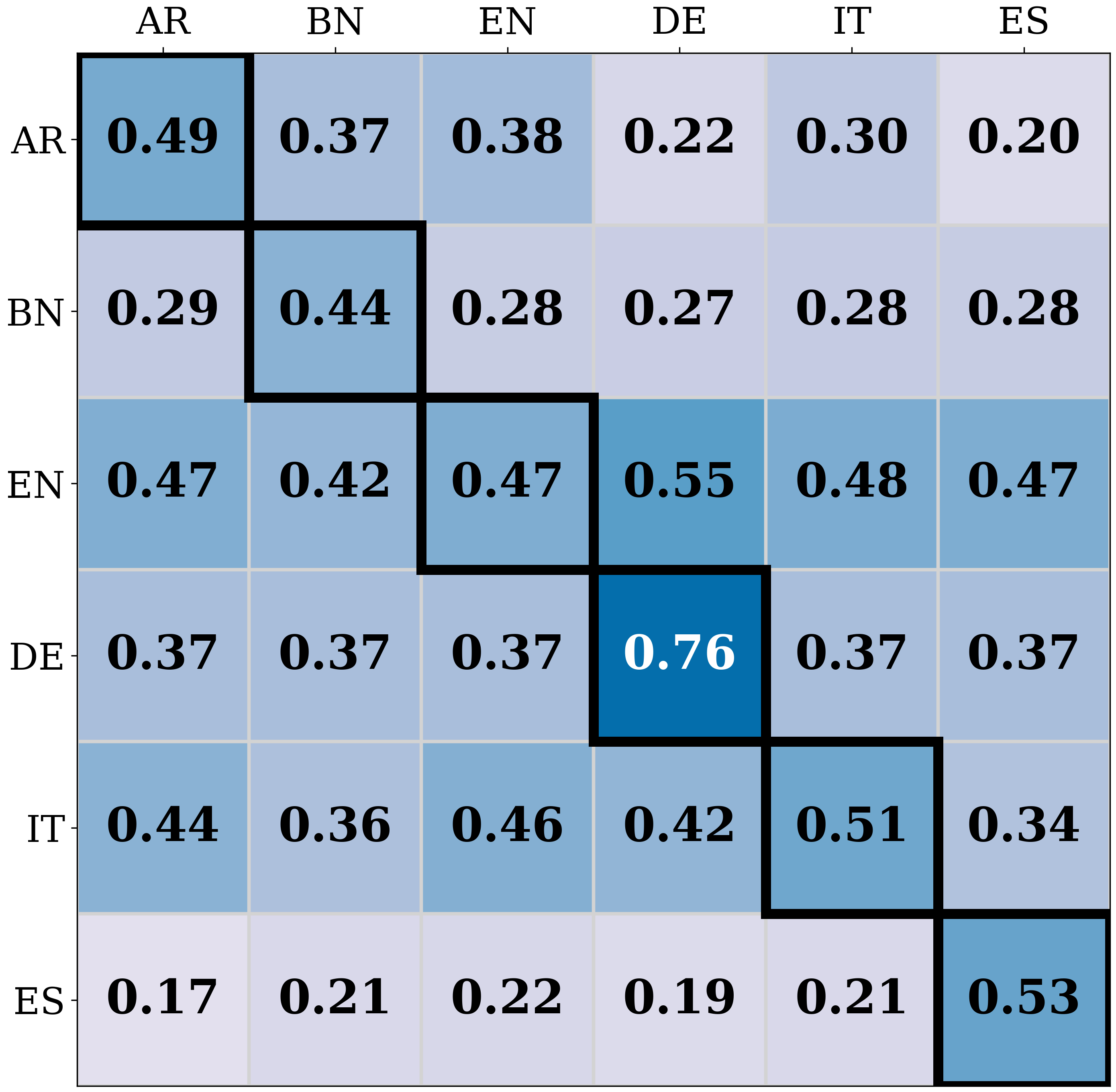}
\vspace{2pt}
{\scriptsize CogVLM2-EN}
\end{minipage} &
\begin{minipage}{0.23\linewidth}
\centering
\includegraphics[width=\linewidth]{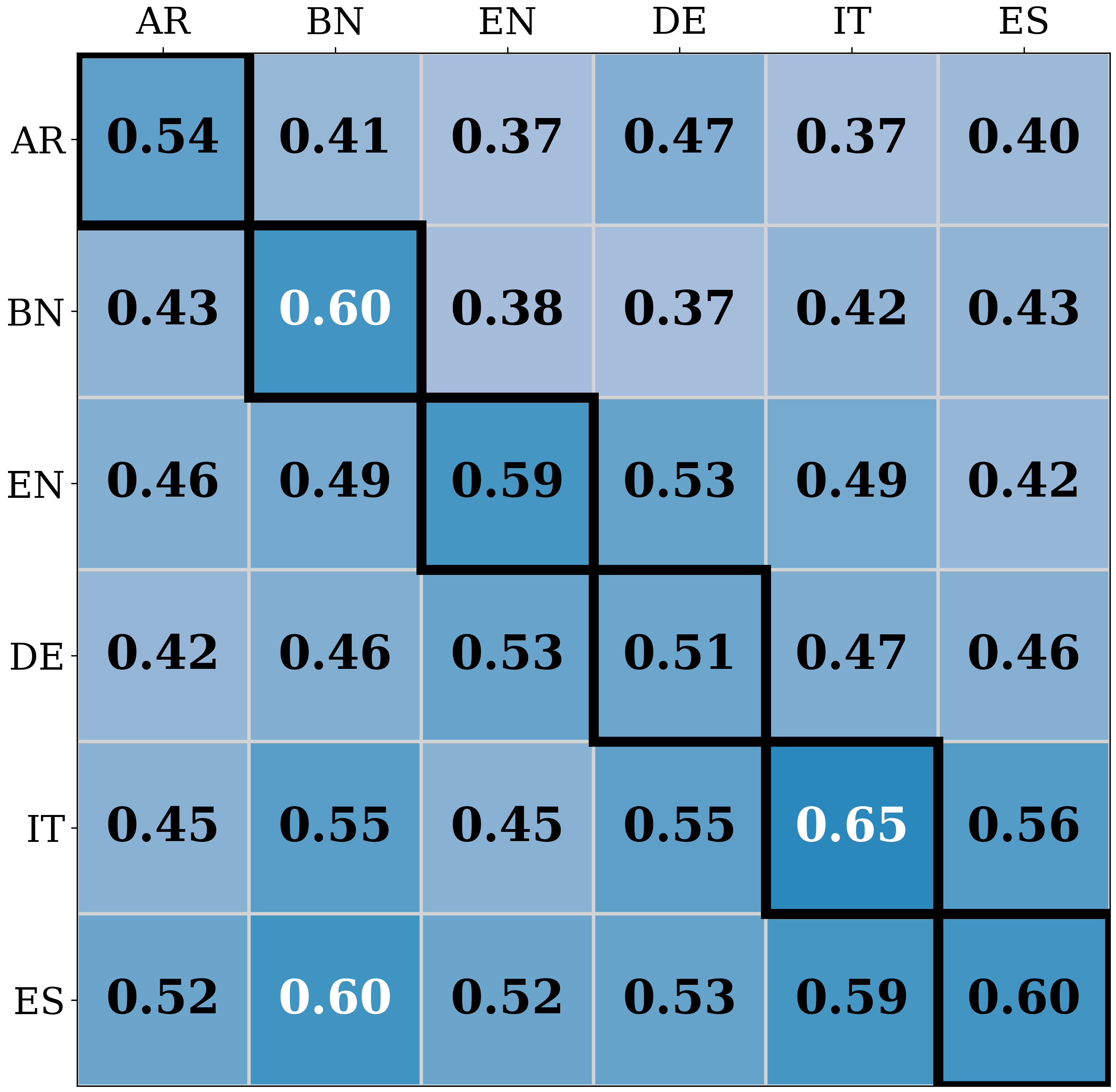}
\vspace{2pt}
{\scriptsize InstructBLIP}
\end{minipage}
\end{tabular}

\vspace{12pt}

\vspace{-1mm}
\Description{Heatmaps for zero-shot performance across seven VLMs.}
\caption{
Heatmaps for zero-shot performance across seven VLMs.
}
\label{fig:heatmap_grid_all_models}
\vspace{-3mm}
\end{figure*}

\section{Analysis and Results}
\label{sec:results}

This section presents empirical results from the hateful meme detection task. The analysis is structured around five major findings that address the research questions, with a focus on the effects of model scale, model specialization, prompting strategies, in-context learning, and translation pipelines on cross-cultural VLM performance.

\subsection{Model Scale and Performance Tiers}

Model scale determines a clear performance hierarchy. However, all models, including state-of-the-art systems, exhibit substantial performance variance across different cultural contexts.
Our analysis first confirms a strong positive correlation between a model's general capabilities (often tied to scale) and its average cross-lingual performance. A distinct tiered hierarchy emerges, with state-of-the-art models like \texttt{gemini-2.5-flash} establishing a top tier of performance, significantly outperforming smaller models on average.

However, a deeper look at performance consistency reveals a more nuanced reality. No model achieves perfect stability. Even the top-performing \texttt{gemini-2.5-flash} shows a wide F1 score distribution, indicating that it struggles to maintain high performance uniformly across all cultural and linguistic contexts. Interestingly, smaller models such as \texttt{instructblip-vicuna-7b} are not simply ``erratic''; rather, they are consistently poor. Their performance distributions are tightly clustered at a low level of effectiveness, indicating that their failure mode in cross-cultural settings is predictable inefficacy, not random variance.

\subsection{General-Purpose vs. Task-Specific Models}

General-purpose VLMs establish a robust cross-lingual baseline, surpassed by task-specific models only in narrowly defined, linguistically proximate transfer scenarios.
We next compare the transferability of general-purpose VLMs against specialized models fine-tuned on a single language. At a macro level, general-purpose models like \texttt{gemini-2.5-flash} and \texttt{gpt-4o-mini} establish a superior average cross-lingual performance baseline. This highlights the critical advantage conferred by broad, multilingual pre-training in establishing a generalized understanding.

\begin{figure}[t]
    \centering
    \begin{subfigure}[t]{0.49\linewidth}
        \centering
        \includegraphics[width=\linewidth]{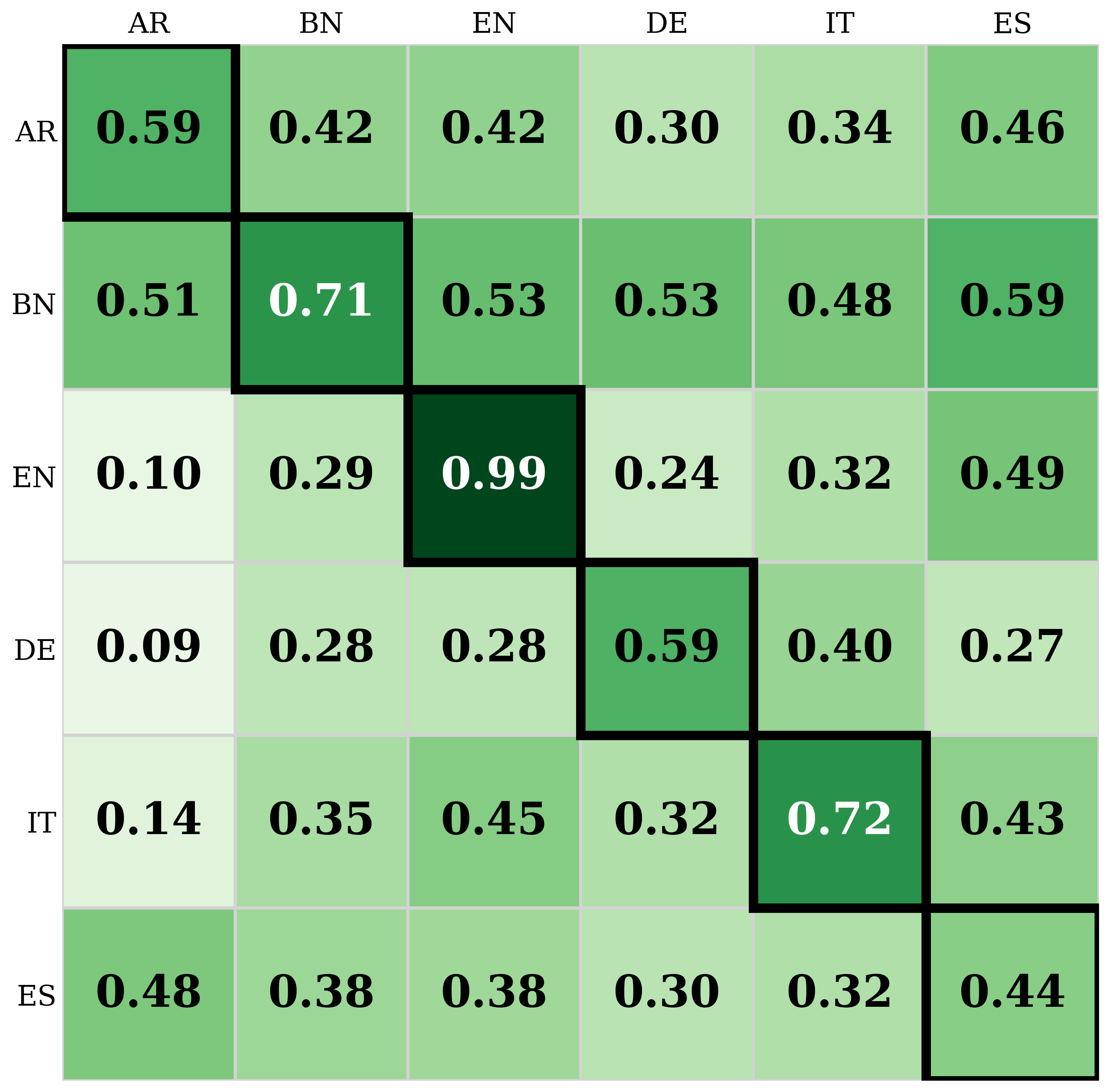}
        \vspace{1pt}
        {\scriptsize Pro-Cap}
    \end{subfigure}\hfill
    \begin{subfigure}[t]{0.49\linewidth}
        \centering
        \includegraphics[width=\linewidth]{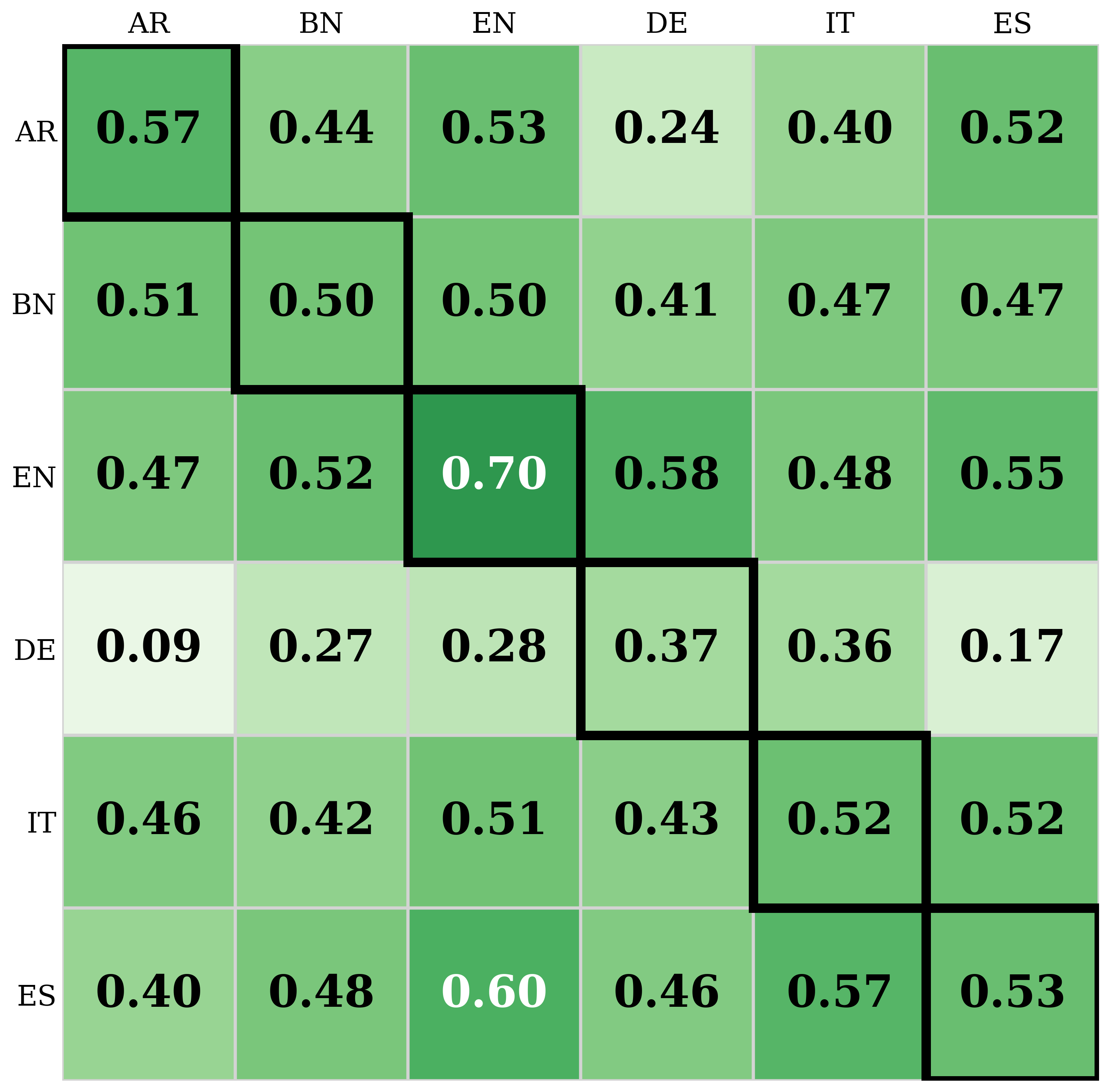}
        \vspace{1pt}
        {\scriptsize PromptHate}
    \end{subfigure}

    \Description{Task-specific transfer heatmaps.}
    \caption{Task-specific transfer heatmaps.}
    \label{fig:task_specific_heatmaps}
    \vspace{-2mm}
\end{figure}

A micro-level analysis using the comprehensive transfer matrix in \Cref{fig:heatmap_grid_all_models} reveals the nature of this trade-off. The rows corresponding to the general-purpose models, particularly \texttt{Gemini (Zero-Shot)}, are uniformly bright, indicating a strong and reliable performance baseline across all target languages. In contrast, the rows for the specialist \texttt{Prompthate} models exhibit a ``feast or famine'' pattern in \Cref{fig:task_specific_heatmaps}: high performance (dark cells) is concentrated in columns where the test language is linguistically proximate to the training language (e.g., the cell for `PromptHate (trained on Spanish)' testing on `italian'). In linguistically distant transfers, their performance collapses (light cells), falling far below the generalist's dependable baseline. This demonstrates the generalist's robustness versus the specialist's narrow, brittle expertise.

\begin{figure*}[t]
\centering
\setlength{\tabcolsep}{2pt}

% ---------------- Row 1: One-shot (4 models) ----------------
% \textbf{One-shot Performance Across Models}\\[3pt]

\begin{tabular}{cccc}
\begin{minipage}{0.23\linewidth}
\centering
\includegraphics[width=\linewidth]{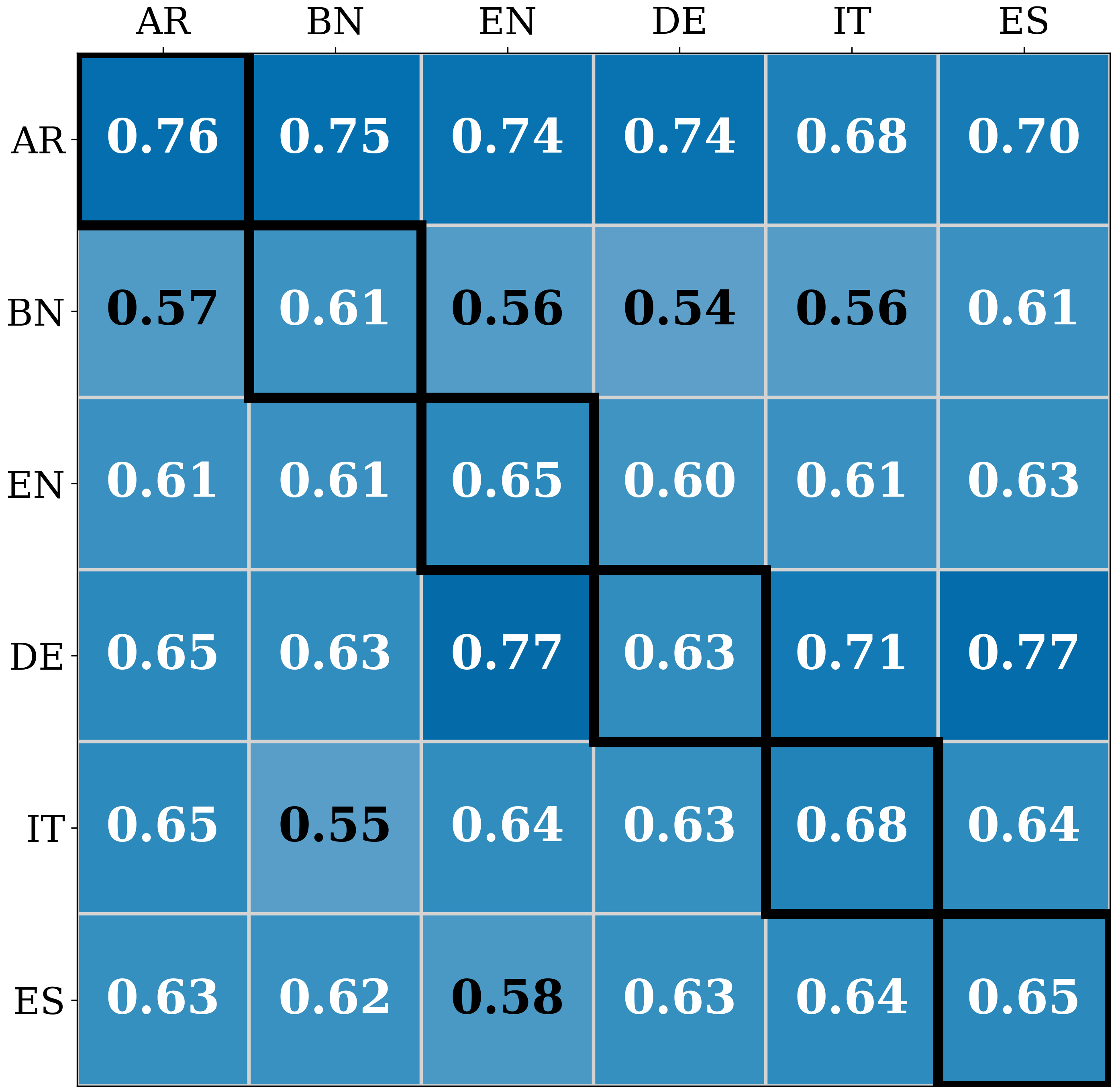}
\vspace{2pt}
{\scriptsize Gemini}
\end{minipage} &
\begin{minipage}{0.23\linewidth}
\centering
\includegraphics[width=\linewidth]{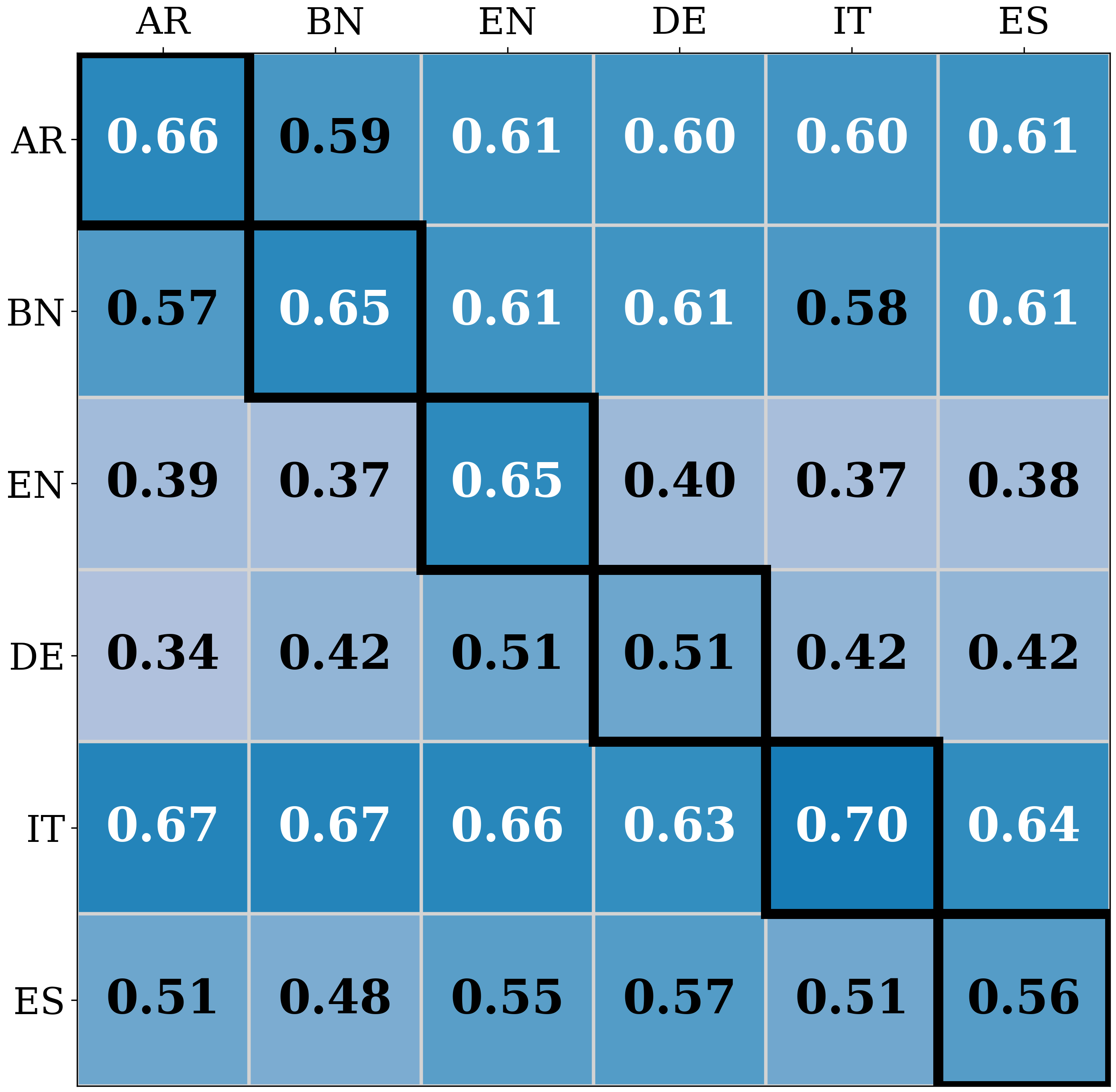}
\vspace{2pt}
{\scriptsize GPT-4o-Mini}
\end{minipage} &
\begin{minipage}{0.23\linewidth}
\centering
\includegraphics[width=\linewidth]{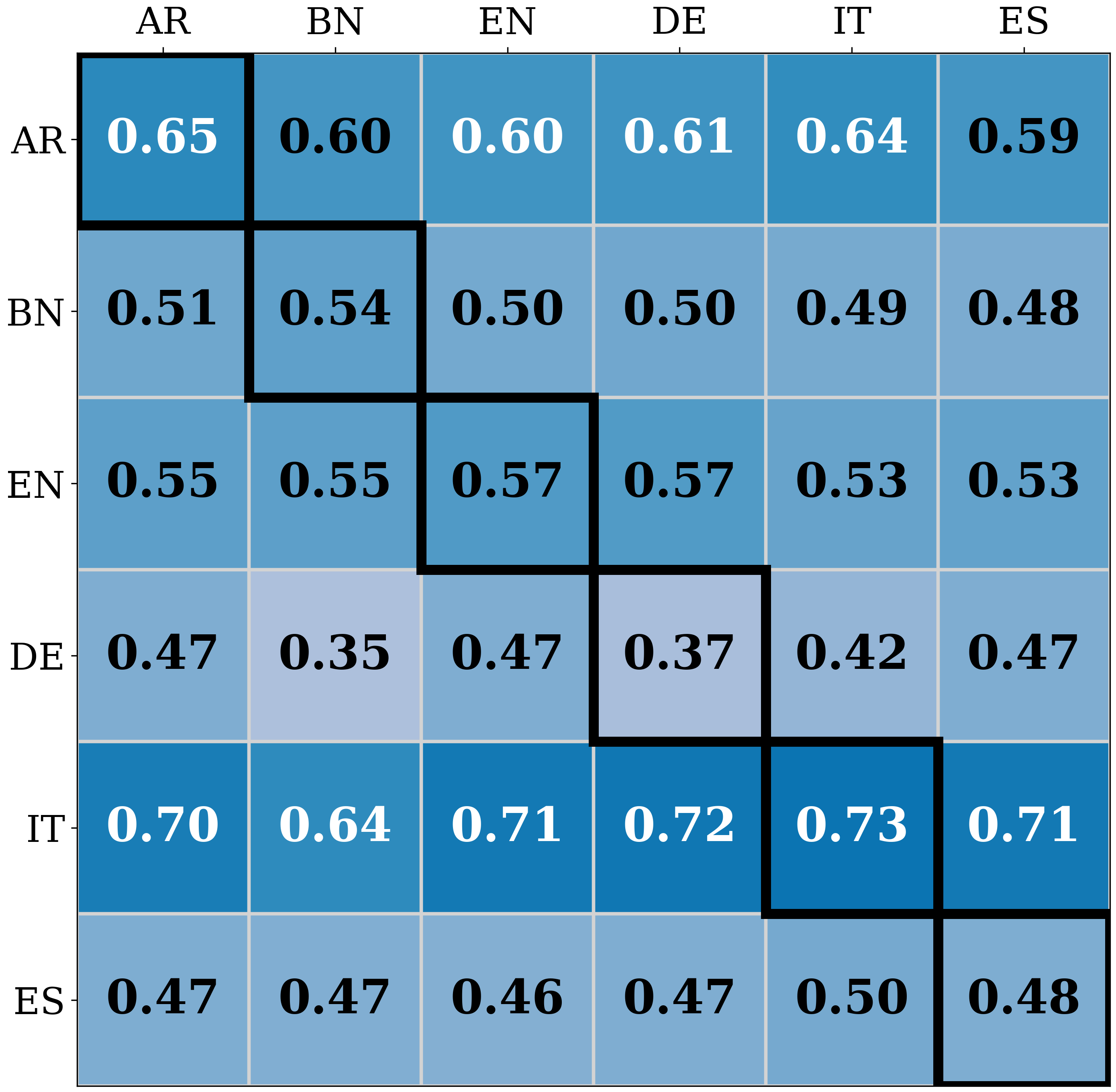}
\vspace{2pt}
{\scriptsize LLaMA-4}
\end{minipage} &
\begin{minipage}{0.23\linewidth}
\centering
\includegraphics[width=\linewidth]{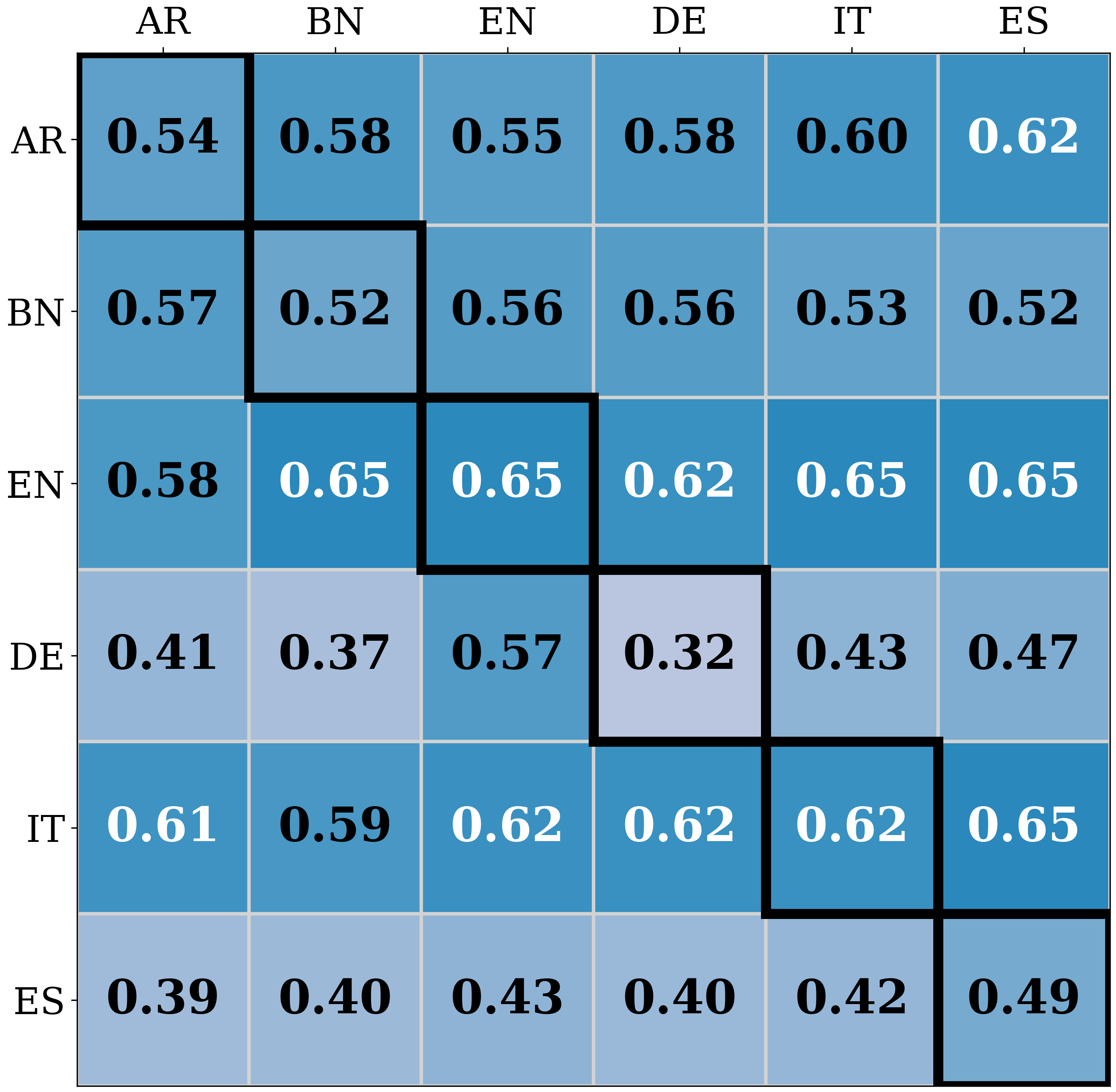}
\vspace{2pt}
{\scriptsize Qwen-VL}
\end{minipage}
\end{tabular}

\vspace{5pt}

% ---------------- Row 2: One-shot (remaining 3 models) ----------------
\begin{tabular}{ccc}
\begin{minipage}{0.23\linewidth}
\centering
\includegraphics[width=\linewidth]{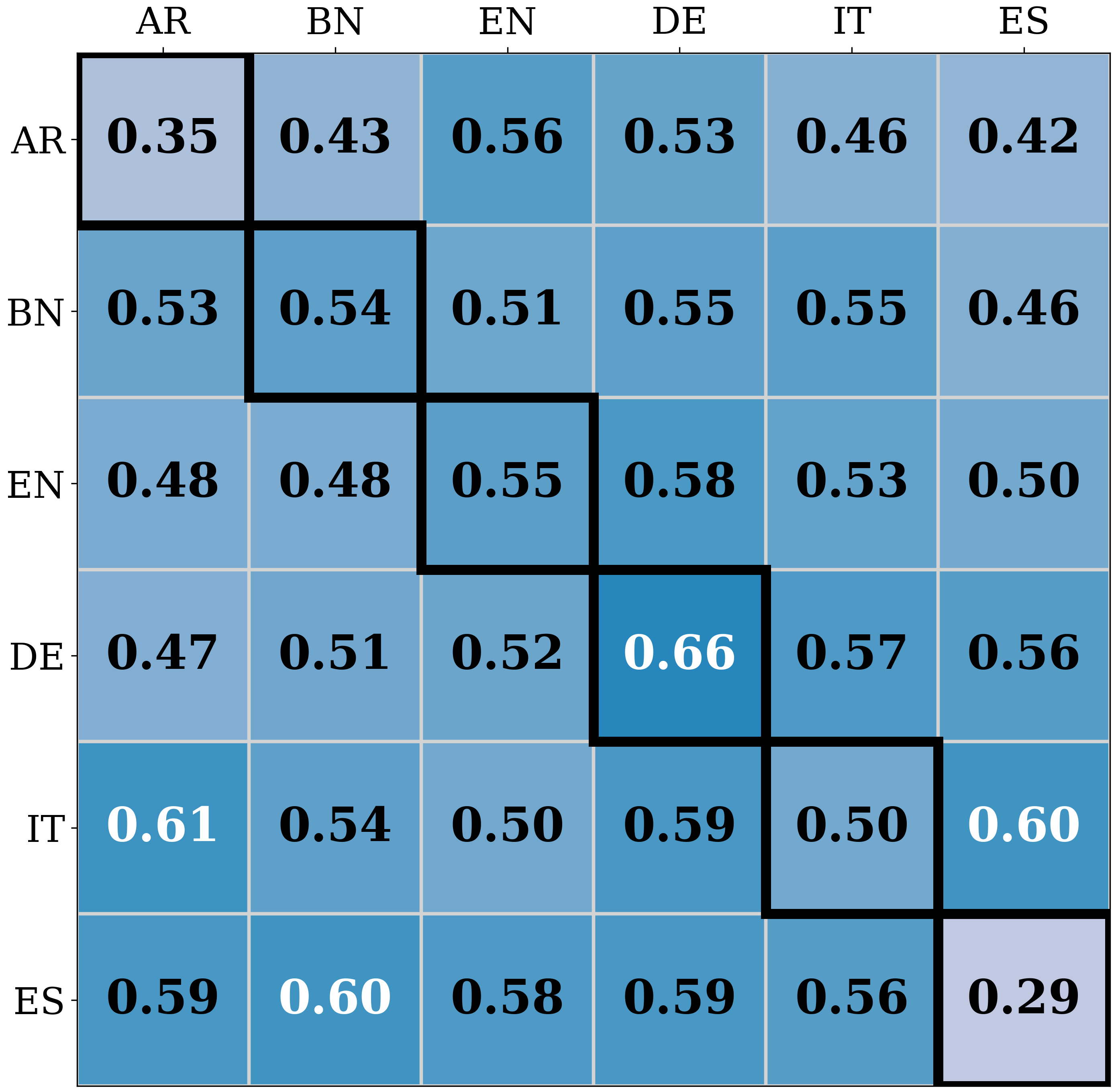}
\vspace{2pt}
{\scriptsize CogVLM2-ZH}
\end{minipage} &
\begin{minipage}{0.23\linewidth}
\centering
\includegraphics[width=\linewidth]{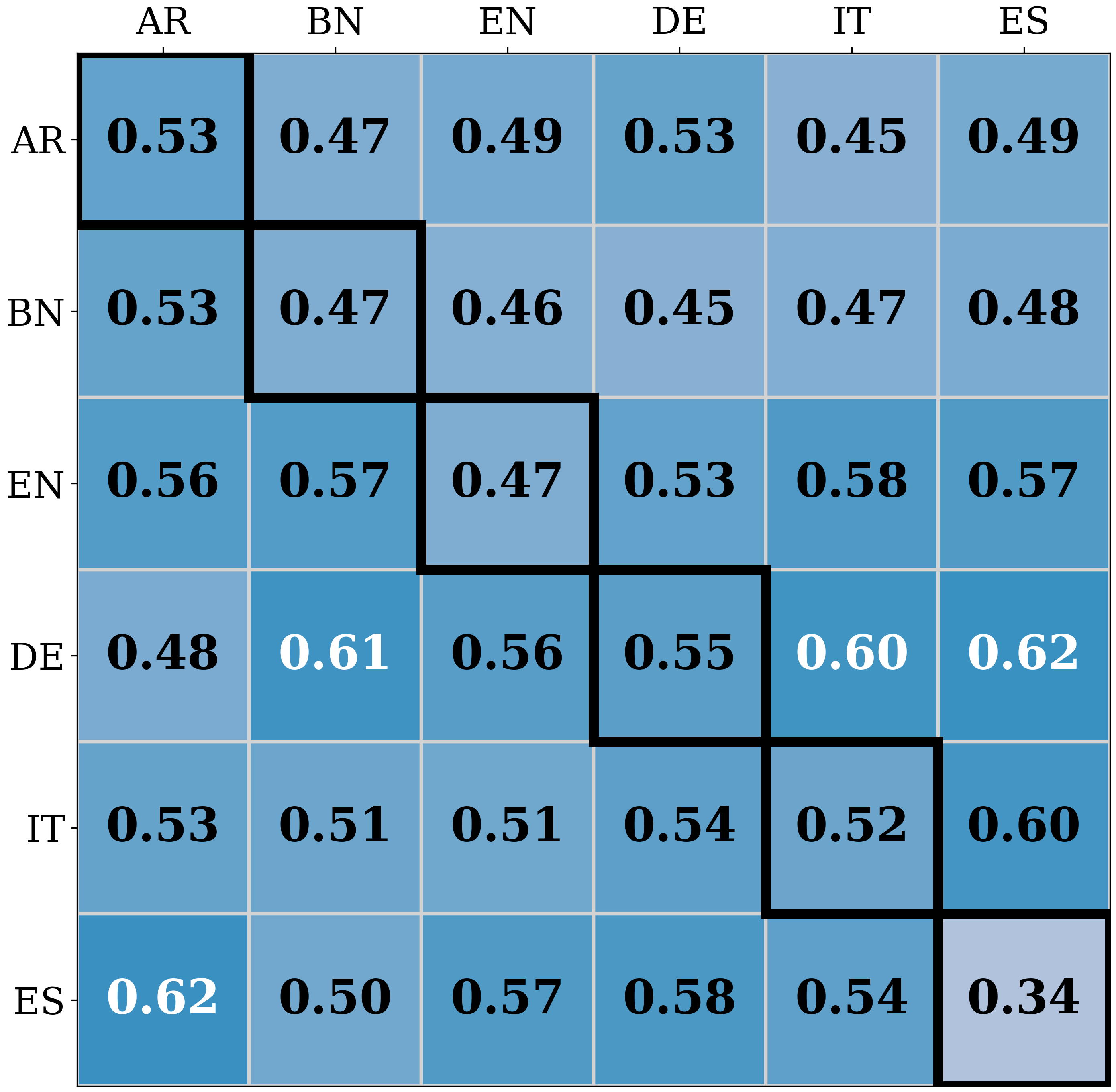}
\vspace{2pt}
{\scriptsize CogVLM2-EN}
\end{minipage} &
\begin{minipage}{0.23\linewidth}
\centering
\includegraphics[width=\linewidth]{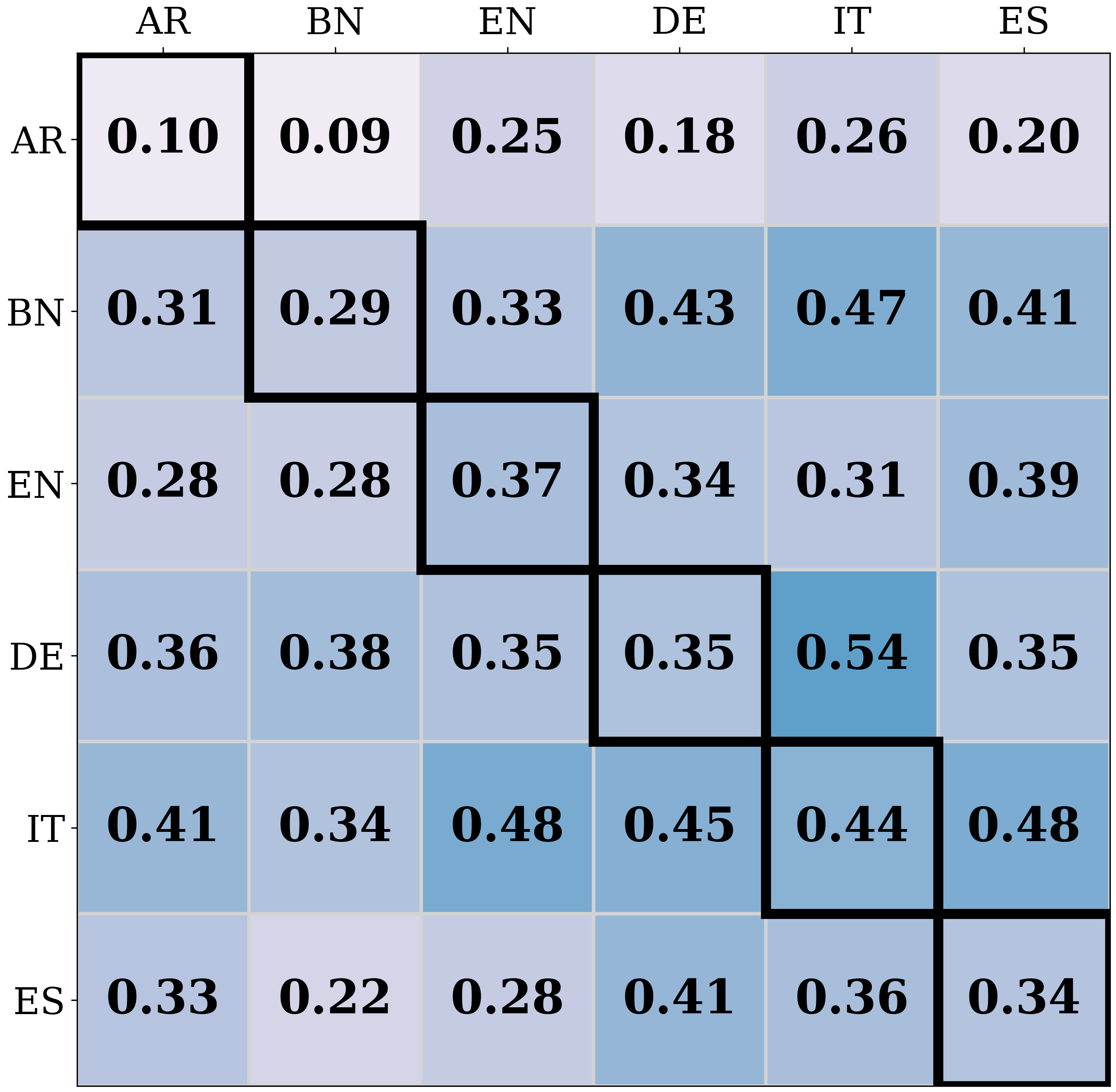}
\vspace{2pt}
{\scriptsize InstructBLIP}
\end{minipage}
\end{tabular}

\Description{Heatmaps One-shot performance across seven VLMs.}
\caption{
Heatmaps One-shot performance across seven VLMs.
}
\label{fig:heatmap_grid_all_models_2}
\vspace{-3mm}
\end{figure*}

\subsection{Prompting Strategies}

Models demonstrate distinct levels of prompt robustness, highlighting the interaction between standardized training and nuanced cultural understanding.
Our analysis of prompting strategies reveals not a simple binary of ``good vs. bad,'' but a clear, three-tiered hierarchy of model robustness. This tiered sensitivity, quantified in the result matrix (\Cref{tab:shot_prompt_models}), provides crucial insights into the internal biases and behavior patterns of different models. Full per-language breakdowns are provided in Appendix \Cref{tab:shot_prompt_all_models}.

\textbf{Tier 1 (Robust Models).} \texttt{gemini-2.5-flash} and \texttt{qwen\allowbreak-2.5\allowbreak-vl\allowbreak-7b\allowbreak-instruct} demonstrate exceptional robustness to prompt language variation. As shown by their low sensitivity scores and the nearly symmetrical performance distributions (\Cref{tab:shot_prompt_models}), their performance remains largely unaffected whether the prompt is in English or a native language. This suggests that these models possess a more generalized, abstract understanding of the hateful content detection task, which is not strictly tied to the linguistic particulars of the prompt.

\textbf{Tier 2 (Moderately Sensitive Models).} \texttt{gpt-4o-mini} and \texttt{llama\allowbreak-4\allowbreak-maverick} form a middle tier of sensitivity. Notably, \texttt{gpt\allowbreak-4o\allowbreak-mini}'s distribution in shows that, on average, native prompts in a zero-shot setting are actually \textit{detrimental} to its performance, while also increasing variance. This counter-intuitive finding suggests a potential interference from its highly tuned, English-centric safety guardrails. Qualitative analysis reveals that native prompts tend to trigger a form of lexical over-sensitivity: the model disproportionately flags content based on specific sensitive keywords in the native language (e.g., slang or aggressive vocabulary) even when the cultural context is benign or sarcastic. 
In contrast, English prompts appear to activate a more abstract, context-aware reasoning pathway, thereby avoiding such false positives. Quantitatively, this manifests as a critical trade-off between over-censorship (False Positive Rate, FPR) and safety misses (False Negative Rate, FNR) (see detailed metrics in the appendix \ref{sec:appendix_safety_analysis}).

\textbf{Tier 3 (Highly Sensitive / Brittle Models).} Smaller models such as \texttt{instructblip-vicuna-7b} and the \texttt{cogvlm} variants are extremely sensitive to prompt language. For \texttt{instructblip}, native prompts act as a critical lifeline, significantly boosting performance. Conversely, for \texttt{cogvlm2-chinese-19b}, native prompts are severely detrimental. This volatility indicates weaker underlying language and cultural understanding, making their performance highly dependent on the phrasing and context of the prompt.

\subsection{In-Context Learning}
One-shot learning serves as an effective cultural intervention, with its impact inversely related to a model's baseline zero-shot performance.

Providing a single, culturally aligned example proves to be a highly effective intervention for improving cross-lingual hateful meme detection. However, the degree of improvement is not uniform across models. As shown in, smaller models exhibit the largest performance gains, with \texttt{instructblip-vicuna-7b} showing the most substantial F1 score increase. By contrast, top-tier models with already strong zero-shot capabilities benefit only marginally from additional examples. contextualizes these gains by illustrating the absolute performance before and after one-shot learning. While weaker models achieve the greatest relative improvements, their final one-shot performance often remains below the zero-shot baseline of larger models. This highlights the role of one-shot learning not as a universal performance booster, but as a targeted mechanism that compensates for knowledge gaps in less capable models.

\subsection{Translation Pipelines}

The translate-to-English strategy consistently degrades detection performance across all models, indicating it is a systematically detrimental approach.
We conclude our analysis by evaluating the common pre-processing pipeline of translating non-English memes into a pivot language (e.g., English) before detection. The performance heatmaps in \Cref{fig:heatmap_grid_all_models} provide a clear and comprehensive visualization of the impact of this strategy.

The results show two consistent patterns. First, across all models and settings, the highest F1 scores are located along the diagonal of each heatmap. These cells represent performance on content in its original, native language, confirming that models perform best when cultural and linguistic context is preserved. Second, the off-diagonal cells, which represent performance on translated captions, exhibit a universal drop in F1 scores. 

This \textit{translation penalty} is pervasive and affects all models, regardless of scale. As shown in \Cref{fig:heatmap_grid_all_models_2}, a comparison between a top-tier model, \texttt{gemini-2.5-flash}, and a smaller model, \texttt{instructblip\allowbreak-vicuna-7b}, reveals that while their baseline performances differ, both are highly susceptible to the information loss inherent in translation. These findings demonstrate that MT strips away essential context that models rely on for accurate detection, making translation-based pipelines an unreliable and counterproductive strategy for global content moderation. Complete heatmaps for all other models are provided in \Cref{sec:appendix_heatmaps}.

\begin{table*}[!t]
\caption{Zero-shot vs.\ one-shot across 6 models (English vs.\ native prompts).}
\label{tab:shot_prompt_models}
\centering
\footnotesize
\setlength{\tabcolsep}{2pt}
\renewcommand{\arraystretch}{1.05}

% ---------- Row 1 ----------
\begin{subtable}[t]{0.31\linewidth}
\centering
\subcaption{gemini-2.5-flash}
\begin{tabular}{@{}l l r r r r r r@{}}
\toprule
& & \multicolumn{2}{c}{ZS} & \multicolumn{2}{c}{OS} & \multicolumn{2}{c}{$\Delta$}\\
\cmidrule(lr){3-4}\cmidrule(lr){5-6}\cmidrule(l){7-8}
Src & Inp. & EN & Nat & EN & Nat & EN & Nat \\
\midrule
Ar & ar & 0.756 & 0.765 & 0.711 & 0.807 & -0.045 & 0.043 \\
Bn & bn & 0.610 & 0.607 & 0.637 & 0.638 & 0.027 & 0.031 \\
En & en & 0.650 & 0.650 & 0.667 & 0.667 & 0.017 & 0.017 \\
De & de & 0.635 & 0.708 & 0.708 & 0.614 & 0.073 & -0.094 \\
It & it & 0.676 & 0.592 & 0.582 & 0.651 & -0.093 & 0.059 \\
Es & es & 0.651 & 0.706 & 0.684 & 0.721 & 0.033 & 0.015 \\
\textit{Avg.} && \textit{0.663} & \textit{0.671} & \textit{0.665} & \textit{0.683} & \textit{0.002} & \textit{0.012} \\
\bottomrule
\end{tabular}
\end{subtable}
\hspace{4pt}
\begin{subtable}[t]{0.31\linewidth}
\centering
\subcaption{gpt-4o-mini}
\begin{tabular}{@{}l l r r r r r r@{}}
\toprule
& & \multicolumn{2}{c}{ZS} & \multicolumn{2}{c}{OS} & \multicolumn{2}{c}{$\Delta$}\\
\cmidrule(lr){3-4}\cmidrule(lr){5-6}\cmidrule(l){7-8}
Src & Inp. & EN & Nat & EN & Nat & EN & Nat \\
\midrule
Ar & ar & 0.655 & 0.316 & 0.734 & 0.716 & 0.079 & 0.400 \\
Bn & bn & 0.654 & 0.604 & 0.609 & 0.565 & -0.045 & -0.038 \\
En & en & 0.646 & 0.646 & 0.726 & 0.726 & 0.081 & 0.081 \\
De & de & 0.513 & 0.490 & 0.513 & 0.591 & 0.000 & 0.100 \\
It & it & 0.700 & 0.737 & 0.649 & 0.683 & -0.051 & -0.054 \\
Es & es & 0.560 & 0.524 & 0.709 & 0.652 & 0.149 & 0.128 \\
\textit{Avg.} && \textit{0.621} & \textit{0.553} & \textit{0.657} & \textit{0.656} & \textit{0.036} & \textit{0.103} \\
\bottomrule
\end{tabular}
\end{subtable}
\hspace{4pt}
\begin{subtable}[t]{0.31\linewidth}
\centering
\subcaption{llama-4-maverick}
\begin{tabular}{@{}l l r r r r r r@{}}
\toprule
& & \multicolumn{2}{c}{ZS} & \multicolumn{2}{c}{OS} & \multicolumn{2}{c}{$\Delta$}\\
\cmidrule(lr){3-4}\cmidrule(lr){5-6}\cmidrule(l){7-8}
Src & Inp. & EN & Nat & EN & Nat & EN & Nat \\
\midrule
Ar & ar & 0.651 & 0.557 & 0.737 & 0.651 & 0.086 & 0.094 \\
Bn & bn & 0.541 & 0.578 & 0.602 & 0.573 & 0.061 & -0.004 \\
En & en & 0.574 & 0.574 & 0.674 & 0.674 & 0.100 & 0.100 \\
De & de & 0.367 & 0.521 & 0.725 & 0.548 & 0.358 & 0.027 \\
It & it & 0.733 & 0.628 & 0.605 & 0.688 & -0.128 & 0.060 \\
Es & es & 0.475 & 0.434 & 0.571 & 0.545 & 0.095 & 0.111 \\
\textit{Avg.} && \textit{0.574} & \textit{0.549} & \textit{0.652} & \textit{0.613} & \textit{0.078} & \textit{0.065} \\
\bottomrule
\end{tabular}
\end{subtable}

\vspace{6pt}

% ---------- Row 2 ----------
\begin{subtable}[t]{0.31\linewidth}
\centering
\subcaption{qwen-2.5-vl-7b-instruct}
\begin{tabular}{@{}l l r r r r r r@{}}
\toprule
& & \multicolumn{2}{c}{ZS} & \multicolumn{2}{c}{OS} & \multicolumn{2}{c}{$\Delta$}\\
\cmidrule(lr){3-4}\cmidrule(lr){5-6}\cmidrule(l){7-8}
Src & Inp. & EN & Nat & EN & Nat & EN & Nat \\
\midrule
Ar & ar & 0.539 & 0.481 & 0.551 & 0.527 & 0.011 & 0.045 \\
Bn & bn & 0.518 & 0.435 & 0.498 & 0.509 & -0.019 & 0.074 \\
En & en & 0.650 & 0.650 & 0.661 & 0.661 & 0.010 & 0.010 \\
De & de & 0.316 & 0.435 & 0.367 & 0.521 & 0.051 & 0.087 \\
It & it & 0.618 & 0.604 & 0.601 & 0.649 & -0.016 & 0.045 \\
Es & es & 0.492 & 0.344 & 0.444 & 0.473 & -0.048 & 0.129 \\
\textit{Avg.} && \textit{0.522} & \textit{0.492} & \textit{0.520} & \textit{0.557} & \textit{-0.003} & \textit{0.065} \\
\bottomrule
\end{tabular}
\end{subtable}
\hspace{4pt}
\begin{subtable}[t]{0.31\linewidth}
\centering
\subcaption{instructblip-vicuna-7b}
\begin{tabular}{@{}l l r r r r r r@{}}
\toprule
& & \multicolumn{2}{c}{ZS} & \multicolumn{2}{c}{OS} & \multicolumn{2}{c}{$\Delta$}\\
\cmidrule(lr){3-4}\cmidrule(lr){5-6}\cmidrule(l){7-8}
Src & Inp. & EN & Nat & EN & Nat & EN & Nat \\
\midrule
Ar & ar & 0.102 & 0.482 & 0.542 & 0.572 & 0.440 & 0.090 \\
Bn & bn & 0.290 & 0.455 & 0.601 & 0.594 & 0.311 & 0.139 \\
En & en & 0.367 & 0.367 & 0.591 & 0.591 & 0.224 & 0.224 \\
De & de & 0.350 & 0.357 & 0.512 & 0.628 & 0.162 & 0.271 \\
It & it & 0.442 & 0.550 & 0.655 & 0.535 & 0.213 & -0.014 \\
Es & es & 0.335 & 0.102 & 0.599 & 0.639 & 0.264 & 0.537 \\
\textit{Avg.} && \textit{0.314} & \textit{0.386} & \textit{0.583} & \textit{0.593} & \textit{0.269} & \textit{0.207} \\
\bottomrule
\end{tabular}
\end{subtable}
\hspace{4pt}
\begin{subtable}[t]{0.31\linewidth}
\centering
\subcaption{cogvlm2-chinese-19b}
\begin{tabular}{@{}l l r r r r r r@{}}
\toprule
& & \multicolumn{2}{c}{ZS} & \multicolumn{2}{c}{OS} & \multicolumn{2}{c}{$\Delta$}\\
\cmidrule(lr){3-4}\cmidrule(lr){5-6}\cmidrule(l){7-8}
Src & Inp. & EN & Nat & EN & Nat & EN & Nat \\
\midrule
Ar & ar & 0.352 & 0.086 & 0.468 & 0.498 & 0.116 & 0.411 \\
Bn & bn & 0.539 & 0.273 & 0.306 & 0.526 & -0.234 & 0.253 \\
En & en & 0.548 & 0.548 & 0.543 & 0.543 & -0.005 & -0.005 \\
De & de & 0.664 & 0.367 & 0.391 & 0.521 & -0.273 & 0.154 \\
It & it & 0.496 & 0.342 & 0.592 & 0.542 & 0.096 & 0.200 \\
Es & es & 0.293 & 0.106 & 0.311 & 0.537 & 0.018 & 0.431 \\
\textit{Avg.} && \textit{0.482} & \textit{0.287} & \textit{0.435} & \textit{0.528} & \textit{-0.047} & \textit{0.241} \\
\bottomrule
\end{tabular}
\end{subtable}

\vspace{-3mm}
\end{table*}

\section{Qualitative Analysis and Cultural Diagnostics}
\label{sec:qualitative_analysis}

To provide interpretability for the quantitative trends, we perform a qualitative review of the model outputs, focusing on cases where native-language prompting caused failure (e.g., \texttt{gpt-4o-mini}) or success (e.g., \texttt{instructblip-vicuna-7b}).

\subsection{Safety-Prior Interference (Tier 2)}
The counter-intuitive performance drop of Tier 2 models (like \texttt{gpt-4o-mini}) in native zero-shot mode is primarily driven by a conflict with its English-centric safety alignment. When the prompt is in English, the model's output converges toward a standardized, risk-averse classification. However, when presented with a native prompt containing unfamiliar colloquialisms or culturally specific references (e.g., political satire in German or Mexican Spanish slang), the model's internal safety system becomes confused. It fails to match the input to its English safety rules, leading to less predictable and often incorrect classifications that are detrimental to performance.

\begin{figure}[!ht]
    \centering
    \includegraphics[width=0.6\columnwidth]{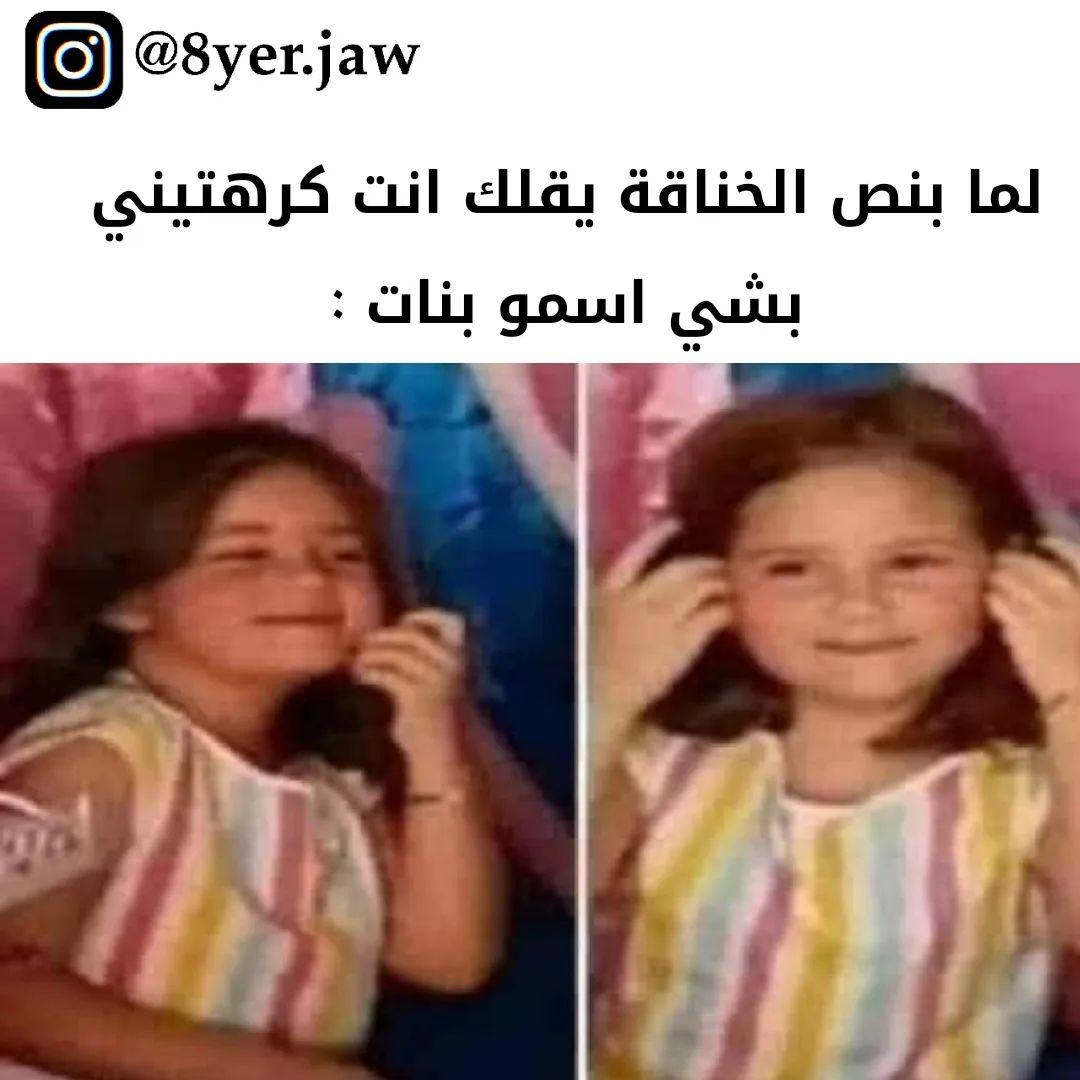}
    \Description{This figure illustrates the corresponding meme sample used in our analysis.}
    \caption{Arabic meme illustrating Semantic Flip failure case.}
    \label{fig:arabic_semantic_flip}
    % \vspace{-0.5mm}
\end{figure}

\paragraph{Case Example: Semantic Flip (Arabic).}
Consider a meme (Figure \ref{fig:arabic_semantic_flip}) collected from Arabic communities, where the image depicts a common relationship trope: a person reacting in exaggerated innocence to a partner's complaint. The text, in Arabic colloquial usage, is an emotional, exaggerated complaint made during an argument, roughly translating to: "You made me hate girls."

In its native context, the phrase is interpreted as harmless, exaggerated frustration (similar to “you’ve scared me off from dating”). The cultural semantic of the verb used is weaker than literal hate. 
However, MT renders this as the literal English phrase: "You made me hate girls."
In the English safety framework, the model’s guardrails are strictly trained to flag the pattern \textnormal{“hate + protected group (gender)”} as hate speech. This literal phrasing triggers a strong response from the model, resulting in a false-positive hate classification. This example illustrates how translation strips away the original pragmatic context (a harmless complaint within a relationship) and forces the multimodal content into a Western safety template, leading to over-moderation and severe loss of cultural fidelity.

\subsection{Cultural Context Loss and Semantic Distortion}
The pervasive ``translation penalty'' (Section 4.5) is explained by the loss of non-literal meaning:
\begin{itemize}
    \item \textbf{Loss of Sarcasm:} Bengali $\rightarrow$ English translations often fail to preserve subtle markers of sarcasm or layered humor common in South Asian memes, flattening them into benign statements that evade detection.
    \item \textbf{Flattening of Dialect:} Arabic $\rightarrow$ English translation often flattens dialectal expressions or region-specific political references into generic, neutral phrases. This strips away the context necessary to trigger a hate classification.
\end{itemize}

\subsection{Representation Gaps (Tier 3)}
The catastrophic collapse of models like \texttt{cogvlm2-chinese-19b} in non-Chinese contexts (e.g., Arabic, German) highlights a severe representation gap. These models are capable only within their primary training domain, confirming that they lack the generalizable cultural visual features needed for multimodal content outside their training ecosystem.

\subsection{Translation Quality Diagnostics}
\label{sec:translation_quality}

To quantify the information loss associated with the "translate-then-detect" pipeline, we conducted a human evaluation. For each language pair (Source $\leftrightarrow$ English), two native speakers independently annotated 50 randomly sampled translated captions on a 4-level scale across two dimensions. The averaged results are summarized below:

The results confirm that while MT generally maintains high linguistic correctness (high average scores), meaning preservation is more variable, particularly for culturally rich content (e.g., Bengali, Spanish). This diagnostic supports our empirical finding: the MT step systematically introduces subtle semantic distortion, which is sufficient to cause the universal performance drop across VLMs.

\begin{table}[h]
\caption{Human Evaluation of Google Translate Quality (4=Best).}
\centering
\footnotesize
\label{tab:translation_quality}
\begin{tabular}{m{25mm} c c}
\toprule
\textbf{Language Pair} & 
\textbf{Correct.} & 
\textbf{Meaning} \\
\midrule
Arabic $\rightarrow$ English & 3.75 & 3.75 \\
Bengali $\rightarrow$ English & 3.00 & 3.00 \\
English $\rightarrow$ Arabic & 4.00 & 4.00 \\
English $\rightarrow$ Bengali & 3.00 & 3.00 \\
English $\rightarrow$ German & 4.00 & 4.00 \\
English $\rightarrow$ Italian & 3.33 & 3.33 \\
German $\rightarrow$ English & 4.00 & 4.00 \\
Italian $\rightarrow$ English & 3.22 & 3.11 \\
Spanish $\rightarrow$ English & 2.89 & 2.89 \\
\bottomrule
\end{tabular}
% \vspace{-2mm}
\end{table}

\section{Conclusion}
\label{sec:conclusion}

We present a systematic, multi-dimensional evaluation of SOTA VLMS for multilingual hateful meme detection, highlighting their strengths and limitations across various languages, prompting strategies, and learning paradigms. Our analysis reveals performance variance and persistent Western-centric bias, even in large models, underscoring the need for culturally aware evaluation. We show that culturally aligned interventions, such as native-language prompting and one-shot learning, effectively mitigate these biases, whereas the common translate-to-English approach reduces detection accuracy. These findings show the importance of moving beyond multilingual benchmarks toward truly multicultural models, and of developing evaluation metrics that account for robustness, fairness, and cross-cultural consistency in addition to accuracy. Limitations and future work are discussed in Appendix~\ref{sec:limitations}.

\section*{Ethics Statement}
\label{sec:ethics}

This research addresses sensitive content and involves several ethical considerations that have been systematically evaluated.

\paragraph{Data Usage.}
All datasets used in this study are publicly available and have been created for research purposes. We do not interact with or collect data from users. In this paper, we have refrained from displaying gratuitously offensive meme examples. The examples included are selected to be illustrative of a specific linguistic or cultural point necessary for the scientific discussion, and they are properly cited. Researchers involved in this study were aware of the potential for exposure to harmful content, and care was taken to minimize prolonged viewing.

\paragraph{Potential for Misuse.}
Research that identifies weaknesses in harmful content detection models carries an inherent risk of being misused by malicious actors to create content that can evade detection. We believe this risk is outweighed by the significant benefit of providing a public, transparent analysis that can help developers and platforms build more robust, culturally-aware, and equitable safety systems. Our findings on the failure of translation pipelines and the benefits of native-context analysis are intended to contribute to more effective global moderation practices directly.

\paragraph{Deployment Considerations.}
Our findings suggest that relying exclusively on ``translate-then-detect'' pipelines for global moderation is risky, as it systematically degrades performance. For practical deployment, we advocate for a hybrid ensemble strategy: rather than trusting a single translation-based model, global moderation systems should ideally cascade locally fine-tuned smaller models (to capture specific cultural nuances and slang) with powerful general-purpose VLMs (to provide broad safety oversight). This approach mitigates the brittleness of translation while balancing the trade-off between cultural sensitivity and general safety guardrails.

\paragraph{Bias and Fairness.}
The central ethical motivation of this work is to investigate and address the cultural biases present in state-of-the-art AI models. Our research demonstrates that current systems tend to align with Western cultural norms, potentially leading to the under-moderation of harmful content in non-Western contexts and the over-moderation of culturally nuanced but benign content. Furthermore, the consistent failure of the “translate-then-detect” pipeline introduces a translation bias, where the loss of culture-specific context can severely compromise the fairness of moderation decisions for non-Western users. By quantifying this bias and testing mitigation strategies, our goal is to foster the development of more equitable AI systems that can serve a global user base fairly and safely. We acknowledge that our own interpretations are shaped by our backgrounds, a common challenge in cultural studies that we have sought to mitigate through objective, data-driven analysis.

\bibliographystyle{ACM-Reference-Format}
\balance
\bibliography{sample-base}

%%
%% If your work has an appendix, this is the place to put it.
\appendix

\section{Prompt Templates}
\label{sec:appendix_prompts}
This section details the prompt structure used in our experiments. We employ a system prompt that defines the task, followed by user prompts containing the meme data. For one-shot experiments, a sequence of user-assistant turns provides in-context examples. The templates are provided in English; for native-language experiments, the corresponding translated version was used.

\paragraph{System Prompt Template.}
\begin{small}
\begin{verbatim}
You are an expert at analyzing images and text. 
You will be shown a meme image and its caption.
Your task is to classify this meme for the task 
of "{task_name}".
Choose one label from the following options: 
{label_options}.

Your response MUST strictly follow this format, 
with no extra text:
reason: [A brief reason for your decision, 
         NO MORE THAN 15 words.]
label: [Choose a single label from the options.]
\end{verbatim}
\end{small}

\paragraph{User Prompt Template for Test Meme.}
\begin{small}
\begin{verbatim}
Caption: "{text}"
\end{verbatim}
\end{small}

\paragraph{One-Shot Example Structure.}
The one-shot examples are inserted between the system prompt and the final user prompt, following this multi-turn format for each example:
\begin{small}
\begin{verbatim}
# User Turn (Example)
Caption: "{example_text}"
[Meme Image]

# Assistant Turn (Example)
reason: {example_reason}
label: {example_label}
\end{verbatim}
\end{small}

\paragraph{One-Shot Example Selection Protocol.}
To ensure the one-shot examples provided in the Native Prompt condition are culturally relevant and unbiased, they were not selected randomly or via retrieval. For each of the six languages, two native speakers, familiar with the corresponding meme culture, independently selected one representative hateful and one non-hateful meme. The final exemplar pair for each language was chosen by consensus between the annotators to be: (1) culturally typical, (2) linguistically natural, and (3) aligned with the dominant meme style. This fixed set of culturally grounded exemplars was used consistently across all models and settings to ensure a fair test of in-context cultural alignment.

\section{Model Details}
\label{sec:appendix_model_details}
To provide a comprehensive benchmark, our evaluation includes two categories of models: state-of-the-art general-purpose VLMs tested in a zero-shot capacity, and specialized models designed specifically for harmful content detection.

\paragraph{General-Purpose VLMs.}
The cross-cultural performance of these models is largely determined by their multilingual training distribution. We assess their intrinsic capabilities based on their reported language coverage:

\begin{itemize}[leftmargin=2mm,topsep=0pt,itemsep=-1pt]
    \item \textbf{Gemini-2.5-Flash}~\citep{comanici2025gemini25}: Gemini models are known for their extensive global multilingual coverage (100+ languages), including major Indo-European languages, Middle Eastern languages (Arabic), South Asian languages (Bengali), and East Asian languages (Chinese). The model utilizes a massive, high-quality multilingual web corpus and cross-modal data. Its consistently stable and high cross-lingual performance in our study aligns with its broad, state-of-the-art multilingual alignment.

    \item \textbf{GPT-4o-Mini}~\citep{openai-2023-gpt4v}: Trained on a diverse multilingual corpus (50+ languages), with a dominant concentration in English and major European languages. While supporting Arabic and some South Asian languages, its system safety framework and semantic alignment rules are inherently English-centric. This explains the instability and performance degradation observed when prompted with non-English native language cues in the zero-shot setting (Tier 2 in Section 4.3).

    \item \textbf{LLaMA-4-Maverick}~\citep{touvron2023llama}: Built upon the LLaMA architectural family, its training data is primarily English-centric, with strong support for major European languages. Support for non-European languages like Arabic, Bengali, and Chinese is significantly weaker (partially multilingual). Its lower and more variable performance on our non-Western datasets, particularly Arabic and Bengali, demonstrates the impact of this training distribution skew.
    
    \item \textbf{Qwen-2.5-VL-7B-Instruct}~\citep{wang2024qwen2vlenhancingvisionlanguagemodels}: The Qwen series is trained with a high proportion of Chinese and English data, but features robust multilingual embedding alignment covering East Asian and a limited set of other languages (including Spanish, Arabic, and French). Its high Prompt Robustness (Tier 1 in Section 4.3) is likely due to the model's strong foundation in cross-lingual alignment despite a skewed distribution.

    \item \textbf{CogVLM2 Family (19B)}: We evaluate both variants to isolate training language effects:
    \begin{itemize}
        \item \textbf{CogVLM2-Chinese-19B}: Trained with a predominantly Chinese corpus and supplementary English data, with minimal coverage of low-resource languages (e.g., Arabic, Bengali). Its catastrophic performance collapse in non-Chinese contexts (Tier 3 in Section 4.3) directly validates the extreme language-specific bias embedded during pre-training.
        \item \textbf{CogVLM2-English-19B}: Primarily trained on English and some European data, exhibiting poor support for non-European languages.
    \end{itemize}

    \item \textbf{InstructBLIP}~\citep{dai2023instructblip}: This VLM leverages instruction tuning on Vicuna-7B. Its performance is often highly dependent on the quality of its instruction-aware visual feature extraction mechanism, but its underlying language foundation is also predominantly English-centric, making it highly sensitive to native prompting in cross-cultural settings.
\end{itemize}

\section{Full Translation Performance Heatmaps}
\label{sec:appendix_heatmaps}

This appendix provides the complete set of performance heatmaps for all general-purpose VLMs evaluated in our study, supplementing the representative examples shown in Section~4.5. Each figure displays the model's F1 score on the hateful meme detection task. The rows indicate the original source language of the meme, while the columns show the language of the caption text presented to the model. The diagonal values represent the performance of the original untranslated content.

\section{Prompting Strategy Full Results}
\label{sec:appendix_detailed_results}

Table~\ref{tab:shot_prompt_all_models} presents the comprehensive breakdown of F1 scores across all six languages and seven models. This table supplements the analysis in Section~\ref{sec:results} by providing the exact numerical values for both English and Native prompting strategies under Zero-shot and One-shot settings.

\section{Detailed Safety Analysis (FPR vs. FNR)}
\label{sec:appendix_safety_analysis}

\textbf{Safety trade-offs: over-censorship vs. missed detection.} While Macro-F1 summarizes overall performance, deployment decisions depend heavily on the specific error profile. As detailed in Table \ref{tab:safety_metrics_appendix}, models exhibit distinct trade-offs between blocking benign content (FPR) and missing harmful content (FNR). 

\noindent Native-language prompting can substantially increase over-censor-ship for some models. For example, GPT-4o-Mini on Arabic shows an FPR increase from 0.22 (ZS-EN) to 0.57 (ZS-NAT), and CogVLM2-English on Arabic reaches 0.39 (ZS-NAT), which is consistent with more keyword-driven rejection under native prompts. In contrast, Gemini-2.5-Flash maintains very low FPR ($<0.05$) but has relatively high FNR in zero-shot settings (e.g., 0.50--0.54 on Arabic), indicating frequent safety misses. Notably, one-shot prompting can mitigate these extremes: for GPT-4o-Mini (Arabic), a single exemplar reduces native-prompt FPR from 0.57 (ZS-NAT) to 0.10 (OS-NAT), improving the safety--utility balance.

\begin{table}[t]
\caption{Safety trade-offs across models and prompting strategies.}
\label{tab:safety_metrics_appendix}
\centering
\tiny
\setlength{\tabcolsep}{1pt}
\renewcommand{\arraystretch}{0.85}

\resizebox{0.75\columnwidth}{!}{%
\begin{tabular}{@{}l l c c c c @{}}
\toprule
\textbf{Lang} & \textbf{Model} & \textbf{ZS-EN} & \textbf{ZS-NAT} & \textbf{OS-EN} & \textbf{OS-NAT} \\
\midrule

% --- Arabic ---
\multirow{7}{*}{\textbf{Ar}}
 & Gemini-2.5-Flash & .04/.50 & .02/.54 & .04/.61 & .04/.33 \\
 & GPT-4o-Mini & .22/.14 & \textcolor{red}{.57/.07} & .10/.32 & .10/.36 \\
 & LLaMA-4-Maverick & .08/.61 & .13/.68 & .08/.36 & .15/.39 \\
 & Qwen-2.5-VL & .20/.54 & .31/.57 & .08/.81 & .14/.79 \\
 & CogVLM2-Chinese-19B & .26/.38 & .30/.61 & .14/.27 & .23/.34 \\
 & CogVLM2-English-19B & .21/.20 & \textcolor{red}{.39/.61} & .22/.26 & .33/.32 \\
 & InstructBLIP & .28/.59 & .14/.24 & .17/.20 & .22/.24 \\
\midrule

% --- Bengali ---
\multirow{7}{*}{\textbf{Bn}}
 & Gemini-2.5-Flash & .06/.70 & .09/.68 & .06/.67 & .22/.50 \\
 & GPT-4o-Mini & .18/.50 & .17/.60 & .09/.69 & .10/.74 \\
 & LLaMA-4-Maverick & .07/.81 & .08/.71 & .10/.67 & .11/.73 \\
 & Qwen-2.5-VL & .20/.73 & \textcolor{red}{.64/.46} & .11/.80 & .39/.56 \\
 & CogVLM2-Chinese-19B & .32/.27 & .34/.45 & .31/.44 & .24/.35 \\
 & CogVLM2-English-19B & .33/.33 & .33/.48 & .28/.34 & .32/.36 \\
 & InstructBLIP & .28/.53 & .22/.35 & .23/.17 & .20/.31 \\
\midrule

% --- English ---
\multirow{7}{*}{\textbf{En}}
 & Gemini-2.5-Flash & .22/.52 & .17/.50 & .23/.48 & .22/.47 \\
 & GPT-4o-Mini & .45/.30 & .29/.36 & .26/.31 & .26/.31 \\
 & LLaMA-4-Maverick & .21/.67 & .21/.68 & .30/.43 & .31/.47 \\
 & Qwen-2.5-VL & .43/.25 & .32/.36 & .16/.56 & .19/.54 \\
 & CogVLM2-Chinese-19B & .20/.32 & \textcolor{red}{.42/.42} & .30/.31 & .24/.26 \\
 & CogVLM2-English-19B & .25/.36 & \textcolor{red}{.37/.44} & .33/.35 & .30/.32 \\
 & InstructBLIP & .32/.40 & .12/.24 & .23/.26 & .23/.26 \\
\midrule

% --- German ---
\multirow{7}{*}{\textbf{De}}
 & Gemini-2.5-Flash & .63/.09 & .50/.09 & .50/.09 & .63/.11 \\
 & GPT-4o-Mini & .75/.18 & \textcolor{red}{.88/.27} & .75/.18 & .63/.18 \\
 & LLaMA-4-Maverick & .50/.73 & .50/.45 & .38/.18 & .63/.27 \\
 & Qwen-2.5-VL & .63/.73 & .75/.36 & .50/.73 & .50/.45 \\
 & CogVLM2-Chinese-19B & .28/.21 & .25/.42 & .37/.40 & .36/.31 \\
 & CogVLM2-English-19B & .28/.24 & .28/.35 & .19/.14 & .20/.22 \\
 & InstructBLIP & .33/.42 & .25/.43 & .31/.38 & .17/.24 \\
\midrule

% --- Italian ---
\multirow{7}{*}{\textbf{It}}
 & Gemini-2.5-Flash & .06/.54 & .08/.65 & .04/.69 & .15/.52 \\
 & GPT-4o-Mini & .25/.35 & .06/.38 & .13/.54 & .13/.48 \\
 & LLaMA-4-Maverick & .06/.44 & .15/.56 & .08/.63 & .17/.44 \\
 & Qwen-2.5-VL & .23/.48 & .39/.40 & .06/.65 & .38/.33 \\
 & CogVLM2-Chinese-19B & .28/.34 & .28/.36 & .25/.21 & .27/.21 \\
 & CogVLM2-English-19B & .33/.26 & .27/.38 & .31/.35 & .24/.28 \\
 & InstructBLIP & .15/.94 & .23/.32 & .19/.18 & .17/.34 \\
\midrule

% --- Spanish ---
\multirow{7}{*}{\textbf{Es}}
 & Gemini-2.5-Flash & .12/.30 & .09/.35 & .09/.41 & .21/.21 \\
 & GPT-4o-Mini & .17/.45 & .11/.54 & .09/.31 & .08/.43 \\
 & LLaMA-4-Maverick & .11/.67 & .06/.73 & .15/.38 & .42/.24 \\
 & Qwen-2.5-VL & .45/.24 & \textcolor{red}{.64/.22} & .41/.33 & .23/.58 \\
 & CogVLM2-Chinese-19B & .41/.32 & .33/.64 & .30/.42 & .30/.26 \\
 & CogVLM2-English-19B & .29/.35 & .31/.32 & .28/.24 & .19/.30 \\
 & InstructBLIP & .14/.85 & .36/.58 & .16/.21 & .21/.24 \\
\bottomrule
\end{tabular}
} % end resizebox

% \vspace{-3mm}
\end{table}

\section{Limitations}
\label{sec:limitations}

Despite the comprehensive analysis, several limitations remain, which present opportunities for future research:

\begin{itemize}[leftmargin=*,topsep=0pt,itemsep=-1pt]
    \item \textbf{Language and Culture Coverage:} Our analysis is based on six language communities. Although diverse, this selection does not represent the full spectrum of global linguistic and cultural diversity. Future work should expand this framework to include more languages, particularly from underrepresented regions such as Africa, Southeast Asia, and indigenous communities.

    \item \textbf{Dataset Representativeness:} We rely on existing public datasets. Although these are valuable resources, they may not fully represent the contemporary or complete digital ecosystem of their respective cultures. The collection methodology and platform-specific nature of these datasets may introduce their own latent biases.

    \item \textbf{Definition of Hate:} Our study uses the labels provided by the source datasets to respect their original, culturally-grounded annotation process. We do not presuppose a universal definition of "hate." Consequently, our findings reflect the models' ability to align with these specific, localized definitions, not an objective, cross-cultural ground truth.

    \item \textbf{Reliance on MT and Translation Noise:} Our experiments involving translated content depend on the quality of the Google Translate API. We conducted a human evaluation of translation quality (summarized in the Discussion), which indicates that while correctness is high, cultural nuances such as sarcasm, local idioms, and dialectal expressions are often lost in the English-pivot translation process. This loss of information is the core mechanism behind the systematic performance degradation that we observe in the “translate-then-detect” pipeline. This limitation, while acknowledged, is a controlled part of our evaluation protocol to assess this prevalent real-world strategy.

    \item \textbf{Robustness of One-Shot Example:} We rely on a single fixed set of culturally representative one-shot exemplars chosen by native speakers. Although this ensures cultural validity and eliminates sampling variability, it does not guarantee robustness against different choices of exemplars. Future work should include a sensitivity analysis in multiple fixed exemplar sets to further confirm the reliability of the one-shot boost, especially for smaller models.

    \item \textbf{Scope of Interventions:} We investigate prompting and one-shot learning as key strategies. However, other powerful intervention techniques, such as culturally-aware fine-tuning, parameter-efficient tuning (e.g., LoRA), or Retrieval-Augmented Generation (RAG) with cultural knowledge bases, were not explored and remain a promising direction for future work.

\end{itemize}

\begin{table*}[!b]
\vspace*{6pt}
\caption{Zero-shot vs. One-shot F1 Scores on Native Content across All VLMs and Prompting Strategies.
\textbf{Key:} \textbf{Bold} indicates the highest F1 score for a given model and language (across all four prompting conditions).}
\label{tab:shot_prompt_all_models}
\centering
\footnotesize
\setlength{\tabcolsep}{3pt}
\renewcommand{\arraystretch}{0.95}

\resizebox{\textwidth}{!}{%
\begin{tabular}{@{}l l c c c c | c c @{}}
\toprule
\textbf{Language} & \textbf{Model} &
\makecell{\textbf{ZeroShot F1}\\ \textbf{(English)}} &
\makecell{\textbf{Native Prompt}\\ \textbf{ZeroShot F1}} &
\makecell{\textbf{OneShot F1}\\ \textbf{(English)}} &
\makecell{\textbf{Native Prompt}\\ \textbf{OneShot F1}} &
\makecell{\textbf{ZS $\Delta$}\\ \textbf{(Nat $-$ En)}} &
\makecell{\textbf{OS $\Delta$}\\ \textbf{(Nat $-$ En)}} \\
\midrule

\multirow{7}{*}{\textbf{Arabic}}
& cogvlm2-chinese-19B & 0.3522 & 0.0863 & 0.4680 & 0.4977 & \textcolor{red}{-0.2659} & 0.0297 \\
& cogvlm2-english-19B & 0.5337 & 0.0863 & 0.4913 & 0.4912 & \textcolor{red}{-0.4474} & \textcolor{red}{-0.0001} \\
& gemini-2.5-flash & \textbf{0.7563} & \textbf{0.7646} & 0.7113 & \textbf{0.8072} & 0.0083 & 0.0959 \\
& gpt4o-mini & 0.6553 & 0.3159 & 0.7342 & 0.7159 & \textcolor{red}{-0.3394} & \textcolor{red}{-0.0183} \\
& llama-4-maverick & 0.6511 & 0.5568 & \textbf{0.7372} & 0.6505 & \textcolor{red}{-0.0943} & \textcolor{red}{-0.0867} \\
& qwen-2.5-vl-7b-instruct & 0.5391 & 0.4814 & 0.5505 & 0.5265 & \textcolor{red}{-0.0577} & \textcolor{red}{-0.0240} \\
\midrule

\multirow{7}{*}{\textbf{Bengali}}
& cogvlm2-chinese-19B & \textbf{0.5393} & 0.2733 & 0.3056 & 0.5261 & \textcolor{red}{-0.2660} & 0.2205 \\
& cogvlm2-english-19B & 0.4740 & 0.2733 & 0.4430 & 0.5130 & \textcolor{red}{-0.2007} & 0.0700 \\
& gemini-2.5-flash & 0.6099 & \textbf{0.6068} & \textbf{0.6371} & \textbf{0.6381} & \textcolor{red}{-0.0031} & 0.0010 \\
& gpt4o-mini & \textbf{0.6543} & 0.6037 & 0.6092 & 0.5653 & \textcolor{red}{-0.0506} & \textcolor{red}{-0.0439} \\
& llama-4-maverick & 0.5414 & 0.5778 & 0.6023 & 0.5735 & 0.0364 & \textcolor{red}{-0.0288} \\
& qwen-2.5-vl-7b-instruct & 0.5178 & 0.3438 & 0.4985 & \textbf{0.5088} & \textcolor{red}{-0.1740} & 0.0103 \\
\midrule

\multirow{7}{*}{\textbf{English}}
& cogvlm2-chinese-19B & 0.5480 & 0.5480 & 0.5428 & 0.5217 & 0.0000 & 0.0000 \\
& cogvlm2-english-19B & 0.4720 & 0.4720 & 0.4700 & 0.5072 & 0.0000 & 0.0000 \\
& gemini-2.5-flash & 0.6500 & 0.6500 & 0.6670 & 0.6670 & 0.0000 & 0.0000 \\
& gpt4o-mini & 0.6455 & 0.6455 & \textbf{0.7262} & \textbf{0.7262} & 0.0000 & 0.0000 \\
& llama-4-maverick & 0.5740 & 0.5740 & 0.6736 & 0.6736 & 0.0000 & 0.0000 \\
& qwen-2.5-vl-7b-instruct & \textbf{0.6504} & \textbf{0.6504} & 0.6607 & 0.6607 & 0.0000 & 0.0000 \\
\midrule

\multirow{7}{*}{\textbf{German}}
& cogvlm2-chinese-19B & \textbf{0.6640} & 0.3667 & 0.3910 & 0.5210 & \textcolor{red}{-0.2973} & 0.1300 \\
& cogvlm2-english-19B & 0.5476 & 0.3667 & \textbf{0.7590} & 0.5778 & \textcolor{red}{-0.1809} & \textcolor{red}{-0.1812} \\
& gemini-2.5-flash & 0.6346 & \textbf{0.7077} & 0.7077 & \textbf{0.6136} & 0.0731 & \textcolor{red}{-0.0941} \\
& gpt4o-mini & 0.5128 & 0.4904 & 0.5128 & 0.5908 & \textcolor{red}{-0.0224} & 0.0780 \\
& llama-4-maverick & 0.3667 & 0.5210 & \textbf{0.7246} & 0.5476 & 0.1543 & \textcolor{red}{-0.1770} \\
& qwen-2.5-vl-7b-instruct & 0.3158 & 0.4345 & 0.3667 & \textbf{0.5210} & 0.1187 & 0.1543 \\
\midrule

\multirow{7}{*}{\textbf{Italian}}
& cogvlm2-chinese-19B & 0.4961 & 0.3421 & 0.5920 & 0.5423 & \textcolor{red}{-0.1540} & \textcolor{red}{-0.0497} \\
& cogvlm2-english-19B & 0.5183 & 0.3421 & 0.5110 & 0.5354 & \textcolor{red}{-0.1762} & 0.0244 \\
& gemini-2.5-flash & 0.6757 & 0.5924 & 0.5824 & \textbf{0.6511} & \textcolor{red}{-0.0833} & 0.0687 \\
& gpt4o-mini & 0.6999 & \textbf{0.7371} & \textbf{0.6486} & 0.6829 & 0.0372 & 0.0343 \\
& llama-4-maverick & \textbf{0.7332} & 0.6279 & 0.6053 & \textbf{0.6875} & \textcolor{red}{-0.1053} & 0.0822 \\
& qwen-2.5-vl-7b-instruct & 0.6176 & 0.6040 & 0.6009 & \textbf{0.6491} & \textcolor{red}{-0.0136} & 0.0482 \\
\midrule

\multirow{7}{*}{\textbf{Spanish}}
& cogvlm2-chinese-19B & 0.2934 & 0.1058 & 0.3113 & 0.5367 & \textcolor{red}{-0.1876} & 0.2254 \\
& cogvlm2-english-19B & 0.3428 & 0.2713 & 0.5250 & \textbf{0.5593} & \textcolor{red}{-0.0715} & 0.0343 \\
& gemini-2.5-flash & 0.6505 & \textbf{0.7060} & 0.6838 & \textbf{0.7210} & 0.0555 & 0.0372 \\
& gpt4o-mini & 0.5599 & 0.5241 & \textbf{0.7088} & 0.6519 & \textcolor{red}{-0.0358} & \textcolor{red}{-0.0569} \\
& llama-4-maverick & 0.4754 & 0.4343 & 0.5705 & 0.5445 & \textcolor{red}{-0.0411} & \textcolor{red}{-0.0260} \\
& qwen-2.5-vl-7b-instruct & 0.4920 & 0.3441 & 0.4739 & 0.4731 & \textcolor{red}{-0.1479} & \textcolor{red}{-0.0008} \\
\midrule

\multirow{7}{*}{\textbf{Overall Mean}}
& cogvlm2-chinese-19B & 0.4827 & 0.2870 & 0.4354 & 0.5280 & \textcolor{red}{-0.1957} & 0.0926 \\
& cogvlm2-english-19B & 0.4812 & 0.3015 & 0.5338 & 0.5242 & \textcolor{red}{-0.1797} & \textcolor{red}{-0.0096} \\
& gemini-2.5-flash & 0.6669 & 0.6713 & 0.6656 & 0.6830 & 0.0044 & 0.0174 \\
& gpt4o-mini & 0.6212 & 0.5529 & 0.6573 & 0.6559 & \textcolor{red}{-0.0683} & \textcolor{red}{-0.0014} \\
& llama-4-maverick & 0.5735 & 0.5489 & 0.6514 & 0.6125 & \textcolor{red}{-0.0246} & \textcolor{red}{-0.0389} \\
& qwen-2.5-vl-7b-instruct & 0.5228 & 0.4916 & 0.5201 & 0.5570 & \textcolor{red}{-0.0312} & 0.0369 \\
\bottomrule
\end{tabular}
} % end resizebox

\vspace{2pt}
{\scriptsize
\setlength{\tabcolsep}{3pt}
\renewcommand{\arraystretch}{1.05}
\begin{tabular}{@{}l p{0.88\textwidth}@{}}
\textbf{Notes.} & \\
ZS $\Delta$ (Nat $-$ En): & Difference between Native and English prompts in ZeroShot mode. \\
OS $\Delta$ (Nat $-$ En): & Difference between Native and English prompts in OneShot mode. \\
\textcolor{red}{Red values}: & Performance degradation when switching from English to Native prompt. \\
\end{tabular}
}
\end{table*}

%=======================

\end{document}